\documentclass[5p]{elsarticle}

\usepackage{amssymb}
\usepackage{amsmath}

\usepackage{booktabs}
\usepackage{hyperref}
\usepackage{xspace}
\newcommand{\eg}{\emph{e.g.}\@\xspace}
\newcommand{\ie}{\emph{i.e.}\@\xspace}
\newcommand{\cf}{\emph{cf.}\@\xspace}
\usepackage{cleveref}
\usepackage{fancybox}  
\usepackage{xcolor}  
\usepackage{multirow}
\newsavebox\CBox
\def\textBF#1{\sbox\CBox{#1}\resizebox{\wd\CBox}{\ht\CBox}{\textbf{#1}}}

\usepackage{hyperref}
\hypersetup{
    colorlinks=true,
    pdfauthor={Name}
    }

\usepackage{cleveref}
\usepackage{graphicx}
\crefname{figure}{Fig.}{Figs.}
\Crefname{figure}{Fig.}{Figs.}
\crefname{table}{Tab.}{Tabs.}
\Crefname{table}{Tab.}{Tabs.}
\crefname{equation}{Eq.}{Eqs.}
\Crefname{equation}{Eq.}{Eqs.}

\newcommand{\change}[1]{{\textcolor{black}{{#1}}}}

\journal{Journal of \LaTeX\ Templates}

\begin{document}

\begin{frontmatter}

\title{Cross-modal feature fusion for robust point cloud registration with ambiguous geometry}

\author[a]{Zhaoyi Wang\corref{cor}}
\ead{zhaoyi.wang@geod.baug.ethz.ch}
\author[a]{Shengyu Huang}
\author[a,b]{Jemil Avers Butt}
\author[c]{Yuanzhou Cai}
\author[a]{Matej Varga}
\author[a]{Andreas Wieser}

\cortext[cor]{Corresponding author}
           
\address[a]{ETH Z\"{u}rich, Institute of Geodesy and Photogrammetry, 8093 Z\"{u}rich, Switzerland}
\address[b]{Atlas optimization GmbH, 8049 Z\"{u}rich, Switzerland}
\address[c]{University of Z\"{u}rich, 8006 Z\"{u}rich, Switzerland}

\begin{abstract}
Point cloud registration has seen significant advancements with the application of deep learning techniques. However, existing approaches often overlook the potential of integrating radiometric information from RGB images. This limitation reduces their effectiveness in aligning point clouds pairs, especially in regions where geometric data alone is insufficient. When used effectively, radiometric information can enhance the registration process by providing context that is missing from purely geometric data.
In this paper, we propose CoFF, a novel \textbf{C}ross-m\textbf{o}dal \textbf{F}eature \textbf{F}usion method that utilizes both point cloud geometry and RGB images for pairwise point cloud registration. \change{Assuming that the co-registration between point clouds and RGB images is available,} CoFF explicitly addresses the challenges where geometric information alone is unclear, such as in regions with symmetric similarity or planar structures, through a two-stage fusion of 3D point cloud features and 2D image features. It incorporates a cross-modal feature fusion module that assigns pixel-wise image features to 3D input point clouds to enhance learned 3D point features, and integrates patch-wise image features with superpoint features to improve the quality of coarse matching. This is followed by a coarse-to-fine matching module that accurately establishes correspondences using the fused features. 
We extensively evaluate CoFF on four common datasets: 3DMatch, 3DLoMatch, IndoorLRS, and the recently released ScanNet++ datasets. In addition, we assess CoFF on specific subset datasets containing geometrically ambiguous cases. Our experimental results demonstrate that CoFF achieves state-of-the-art registration performance across all benchmarks, including remarkable registration recalls of 95.9\% and 81.6\% on the widely-used 3DMatch and 3DLoMatch datasets, respectively. CoFF is particularly effective in scenarios with challenging geometric features, provided that RGB images are available and that the overlapping regions exhibit sufficient texture in the RGB images. \change{Our code is available at \href{https://github.com/zhaoyiww/CoFF}{https://github.com/zhaoyiww/CoFF}.}

\end{abstract}


\end{frontmatter}


\section{Introduction}
\label{sec:intro}

Pairwise point cloud registration is the task of estimating a relative transformation that aligns two given point clouds. It is important for a wide range of applications like 3D reconstruction~\cite{ZHU202026, elhashash2022, liu_deep_2023}, place recognition~\cite{SHI2023637, XIE202415}, Simultaneous Localization and Mapping (SLAM)~\cite{slam2013,SHAO2020214}, and autonomous driving~\cite{huang2021comprehensive,huang2022accumulation}. Deep learning-based point cloud registration~\cite{ao2023buffer,yu2023rotation,SIRA_PCR_2023_ICCV,Rethinking_2023_ICCV} has recently seen significant progress, particularly for challenging scenarios with low overlap between two point clouds~\cite{predator2021,yu2021cofinet,peal_2023_CVPR}. 
These approaches typically involve learning point feature descriptors, establishing correspondences through feature matching, and estimating transformations using \eg, RANSAC~\cite{81ransac} or the Kabsch algorithm based on Singular Value Decomposition (SVD)~\cite{kabsch1976solution,stewart1993svd}. To improve keypoint saliency and matching efficiency, some methods rely on keypoint detection~\cite{predator2021, bai2020d3feat}. For instance, D3Feat~\cite{bai2020d3feat} combines feature description with keypoint detection, while Predator~\cite{predator2021} introduces overlap and saliency score predictions to detect keypoints.
However, the precision of keypoint detection is often insufficient,
and thus compromises the final registration quality~\cite{yu2021cofinet}.

\begin{figure*}[tb]
  \centering
  \includegraphics[width=1.0\linewidth]{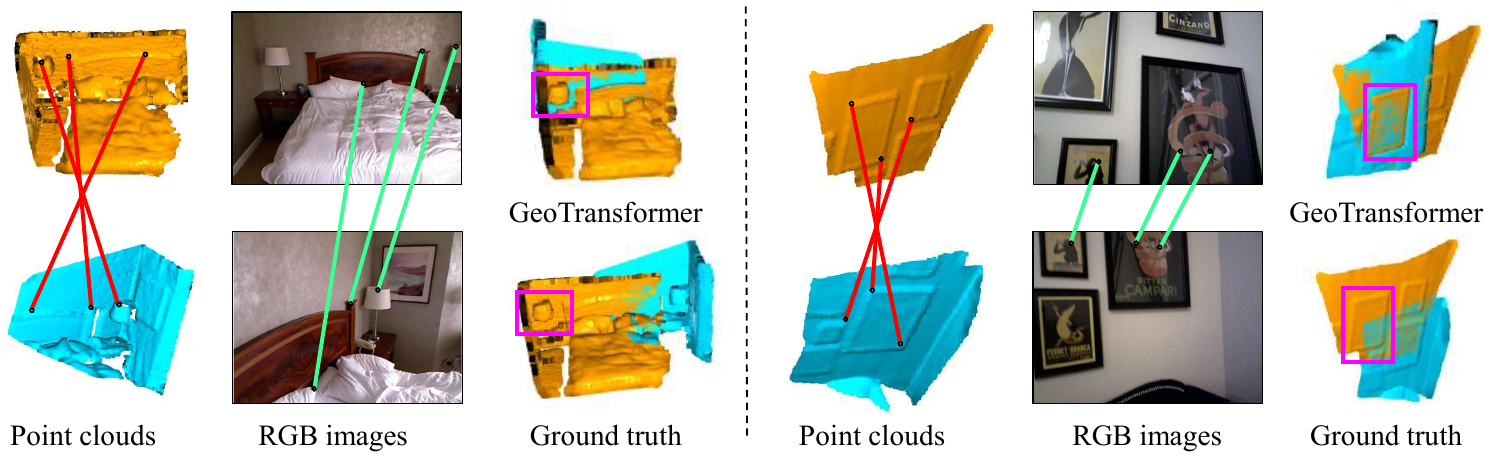}
  \caption{Challenges for PCR due to ambiguous geometry ({\setlength{\fboxsep}{0pt}\colorbox{green}{correct}}, {\setlength{\fboxsep}{0pt}\colorbox{red}{wrong}} correspondences). Left: example from the 3DMatch~\cite{zeng20163dmatch} dataset containing two lamps wrongly matched due to geometric similarity. Right: example from the IndoorLRS~\cite{Park2017coloricp} dataset, showing the overlap of point clouds with extended planar regions. While color information in the associated images would provide valuable clues, the purely geometry-based GeoTransformer~\cite{qin2022geometric} fails to properly align the point clouds.}
  \label{fig:ambiguity}
\end{figure*}

\change{Detector-free} methods, such as those adopting a coarse-to-fine matching strategy~\cite{yu2021cofinet,yang2022oif-pcr,qin2022geometric,yu2023rotation}, have achieved superior performance. This success is largely due to their integration of geometric features not only in feature description but also in the matching process~\cite{yu2021cofinet,qin2022geometric}. For instance, in GeoTransformer~\cite{qin2022geometric}, coarse matching between superpoints (which represent the coarsest level of point aggregates) ensures global consistency. Fine matching then utilizes these superpoints and their corresponding neighborhoods to exploit local geometric details, enhancing matching precision.
\change{In such cases, superpoints offer advantages over keypoints. Unlike keypoints detected by \change{detector-based} methods, which may be unevenly distributed across a scene, superpoints are obtained through hierarchical subsampling and thus maintain a more uniform distribution while preserving structural information.}
However, in scenarios dominated by symmetries or extended planar subsets (\cf \cref{fig:ambiguity}), purely geometry-based \change{detector-free} registration methods struggle to establish correct correspondences and thus fail to align the point clouds. We denote the underlying lack of geometric saliency as \textit{ambiguous geometry} herein.

Learning features from both RGB images and point clouds is known as 2D-3D cross-modal representation learning~\cite{representlearning}. Despite its application to areas such as object detection, segmentation, and scene understanding~\cite{pham2020lcd,ao2020SpinNet,Peng2023OpenScene,dai20183dmv,qi2020imvotenet,logonet,hou2021pri3d,yan2022let,robert2022dva}, applications to point cloud registration have been less explored. One of the few exceptions 
is Color-ICP~\cite{Park2017coloricp} which utilizes both point cloud and image information for registration. However, it requires special initialization schemes due to its vulnerability to local minima. 
LCD~\cite{pham2020lcd} utilizes 2D and 3D auto-encoders to learn cross-modal feature descriptors. 
Despite its promising results, dense patch construction proves computationally expensive. 
Moreover, LCD processes each point cloud individually and does not incorporate information from other point clouds in the pair. It thus leads to relatively weak registration performance. 
More recently, PCR-CG~\cite{zhang2022pcr} employs image features extracted from a 2D pre-trained model~\cite{resnet,hou2021pri3d} to enhance 3D feature description. It inherits the limitation of its 3D backbone Predator~\cite{predator2021}, which has low keypoint detection precision. Furthermore, directly integrating pixel-wise image features into point clouds before passing them to a 3D network, as done by PCR-CG, may not fully exploit the \change{local detail} information of the images.

\begin{figure*}[tb]
  \centering
  \includegraphics[width=1.0\linewidth]{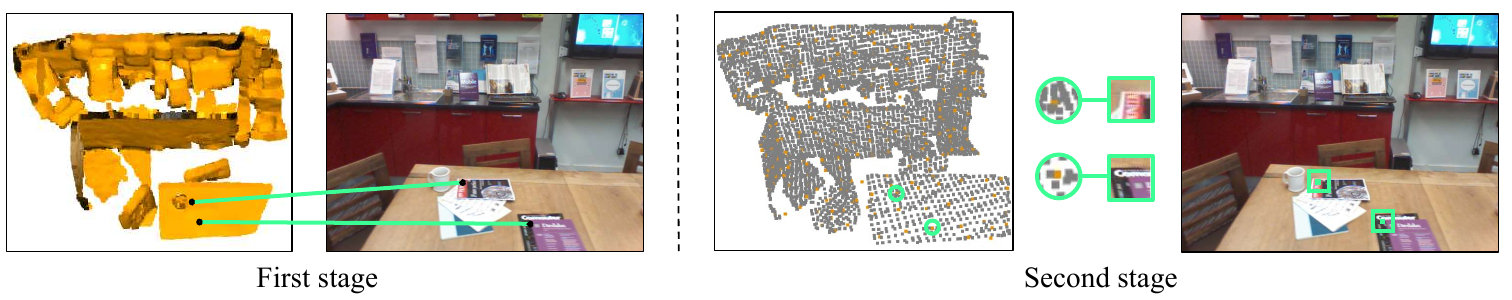}
  \caption{Overview of our two-stage feature fusion process. In the first stage, pixel-wise image features are extracted using a pre-trained fully convolutional image network, and then fused with 3D points in the input point cloud; in the second stage, patch-wise image features are computed with a patch-based image network, and then fused with superpoint features.}
  \label{fig:fusion_overview}
\end{figure*}

To deal with the challenges of ambiguous geometry, and also to fully leverage RGB images, we propose a cross-modal feature fusion method for pairwise point cloud registration. It involves a two-stage process, fusing (i) 3D input point clouds with pixel-wise image features and (ii) superpoint features with patch-wise image features (\cf \cref{fig:fusion_overview}).
The first stage enhances the learned 3D features \change{by incorporating pixel-wise image features, which capture global information across the entire image. These image features are assigned to each point in point clouds through pixel-to-point correspondences}. Meanwhile, the second stage helps alleviate ambiguity caused by geometry during the coarse matching phase \change{by leveraging patch-wise image features, which effectively preserve local details. Therefore, pixel-wise and patch-wise image features are complementary rather than redundant, each playing a distinct role.} More specifically, our method comprises five components: a point feature extractor, a pixel-wise image feature extractor, a patch-wise image feature extractor, a cross-modal feature fusion module, and a coarse-to-fine matching module. The point feature extractor utilizes the widely-used 3D backbone, KPConv-FPN~\cite{thomas2019KPConv,Lin2016FPN}, to extract both superpoint features and dense point features. The pixel-wise image feature extractor employs a 2D fully convolutional network to extract image features. Similarly, the patch-wise image feature extractor extracts per-patch image features, each associated with a superpoint and its corresponding neighborhood.
The cross-modal feature fusion module assigns pixel-wise image features to 3D input point clouds and integrates patch-wise image features with superpoint features. Based on the fused features, a coarse-to-fine matching strategy~\cite{qin2022geometric} is then adopted to establish correspondences, and the final transformation is estimated using a Local-to-Global Registration (LGR) estimator~\cite{qin2022geometric}. 
In summary, our main contributions are as follows:

\begin{itemize}
\item \change{We introduce CoFF, a two stage, cross-modal feature fusion method for point cloud registration. The first stage assigns pixel-wise image features to 3D point clouds, enhancing the point cloud features with image-derived global information. The second stage fuses patch-wise image features, which retain local texture details, with superpoint features, reducing ambiguity caused by geometry in coarse matching.}
\sloppy \item \change{We critically analyze the limitations of existing geometry-based methods in point cloud registration, particularly their performance when faced with real-world challenges such as geometric ambiguity. 
We propose a solution by introducing subset datasets designed for scenarios where these geometric ambiguities pose significant challenges. 
}
\end{itemize}

The remainder of this paper is organized as follows: 
Section \ref{sec:related} offers a literature review and discussion of related previous studies. 
Section \ref{sec:method} describes the principles of our proposed method. Section \ref{sec:experi} presents the datasets, evaluation setup, and experimental results, while Section \ref{sec:discuss} gives a detailed discussion. Finally, the conclusion is outlined in Section \ref{sec:conclusion}.
\section{Related work}
\label{sec:related}

In this section, we review previous studies on deep learning-based point cloud registration methods, covering local patch-based methods, fully convolutional methods, and recent patch-enhanced fully convolutional methods. We also summarize recent advancements in 2D-3D cross-modal representation learning \change{and highlight the differences between our cross-modal feature fusion method to exiting approaches.}

\subsection{Deep learning-based registration}

\subsubsection{\change{Local patch-based methods}}

\change{Local patch}-based registration methods learn feature descriptors from local patches  (\ie, neighborhoods), typically requiring point cloud patches as input. 3DMatch~\cite{zeng20163dmatch} is a pioneer work of these patch-based methods. It is built upon a Siamese 3D network~\cite{Siamese93} that extracts local feature descriptors from a truncated distance function embedding. PPFNet~\cite{deng2018ppf} directly processes 3D points by integrating them with the PointNet~\cite{qi2017pointnet} architecture, enabling the learned descriptors to incorporate a global context. While PPF-FoldNet~\cite{ppffoldenet}, a self-supervised version of PPFNet~\cite{deng2018ppf}, further incorporates the FoldingNet~\cite{yang2018foldingnet} architecture to learn rotation-invariant descriptors. \sloppy 3DSmoothNet~\cite{gojcic20193DSmoothNet} introduces a smoothed density value representation for learning local feature descriptors. SpinNet~\cite{ao2020SpinNet} utilizes a local reference frame to align local patches, and employs 3D cylindrical convolution layers to learn feature descriptors. These patch-based registration methods demonstrate strong generalization ability to unseen datasets, as the local features they learn are applicable across different scenes. However, they often encounter problems such as low computational efficiency and a limited receptive field. This is primarily due to intermediate networks process adjacent patches independently, without leveraging the learned features from one patch to benefit others.

\subsubsection{Fully convolutional methods}
\label{subsec:fully_conv}

To alleviate the problems with \change{local} patch-based methods, fully convolutional registration methods~\change{specifically use fully convolutional networks (\eg, Minkowski~\cite{choy20194mink} and KPConv~\cite{thomas2019KPConv}) that incorporate hierarchical convolutional mechanisms to process the point cloud globally. This differs from the convolutional operations used in local patch-based registration methods, \eg, shared MLPs and pooling.} FCGF~\cite{FCGF2019} is the first method that attempts using a sparse tensor-based fully convolutional network~\cite{choy20194mink} for registration.
FCGF achieves efficient training while it still performs comparably to previous patch-based methods~\cite{zeng20163dmatch,Deng2018PPFNetGC,ppffoldenet,gojcic20193DSmoothNet}. 
Building on a kernel point-based fully convolutional network (KPConv~\cite{thomas2019KPConv}), and also inspired by joint feature detection and description in 2D~\cite{d2net}, D3Feat~\cite{bai2020d3feat} introduces a joint feature description and detection approach for 3D point cloud registration. 
Predator~\cite{predator2021} is the first work that addresses the problem of low overlap, by predicting overlap and saliency scores, and selecting points with high scores as keypoints. However, for \change{detector-based} methods, the precision of these detected keypoints remains an issue, as it limits the final registration precision. 
Compared to patch-wise registration methods, fully convolutional registration methods have larger receptive fields. This allows them to capture features at a global scale. However, this advantage can lead to the loss of local fine-grained details due to hierarchical downsampling operations. Consequently, the loss of these local details can limit the potential for improving registration performance.

\subsubsection{Patch-enhanced fully convolutional methods}

To further improve the performance of fully convolutional registration methods, 
some approaches incorporate local patch-based components. 
For instance, RoReg~\cite{wang2023roreg} utilizes a full-convolutional network while also learning a local patch-based orientation-sensitive descriptor that encodes the anisotropy of local patches. Consequently, this approach effectively combines rotation invariance and rotation equivariance, enhancing the discriminative power of learned feature descriptors. Some 2D image matching techniques employ a coarse-to-fine mechanism to address the repeatability issue in keypoint detection~\cite{li20dualrc,ZhouCVPRpatch2pix,sun2021loftr}. Driven by this, CoFiNet~\cite{yu2021cofinet} introduces a \change{detector-free} coarse-to-fine approach for 3D point cloud registration. Although the features are based on a fully convolutional network, the deep exploration of local patches in the coarse-to-fine matching can considerably boost the registration performance. This strategy has been widely adopted in recent point cloud registration approaches~\cite{SIRA_PCR_2023_ICCV,yu2024learning,mu2024colorpcr}.
GeoTransformer~\cite{qin2022geometric} employs a RANSAC-free LGR strategy. Local registration involves using superpoint correspondences to generate coarse transformation candidates. Similar to CoFiNet, superpoint correspondences enhanced by local patches are decoded to dense point correspondences. The global registration refines (re-estimates) the transformations iteratively using high-quality dense point correspondences.
Buffer \cite{ao2023buffer} benefits from the robust characteristics of local patches, which are inherently resilient to occlusion and easily discriminable. It incorporates both fully convolutional~\cite{thomas2019KPConv} and patch-based~\cite{ao2020SpinNet} modules for keypoint detection and learning feature descriptors, respectively, thereby ensuring a balanced trade-off between accuracy, efficiency and generalizability.

To alleviate the computational burden associated with patch-based components, RoReg~\cite{wang2023roreg} limits the number of points per input point cloud (\eg, 5,000 points). GeoTransformer~\cite{qin2022geometric} constructs patches exclusively  for superpoints, and Buffer~\cite{ao2023buffer} initially employs a point-wise method for keypoint detection, followed by a patch-wise approach to learn descriptors for these keypoints. Although these methods perform strongly by fully exploiting geometric information inherent in 3D point clouds, their reliance solely on geometric cues may be limiting.

\subsection{2D-3D cross-modal representation learning}

The integration of point clouds and RGB images has been the focus of several studies.
One line of research focuses on joint feature representation learning. 
LCD~\cite{pham2020lcd} introduces auto-encoders that map 2D images and 3D point clouds into joint representations. It leverages a triplet loss to reinforce the interaction between 2D and 3D features. PiMAE~\cite{chen2023pimae} employs a pre-training MAE~\cite{MAE2022} with 2D images and 3D point clouds for 3D object detection.
OpenScene~\cite{Peng2023OpenScene} proposes 2D-3D joint learning for open vocabulary 3D scene understanding in which inference relies partly on CLIP~\cite{Radford2021clip} text features. The cross-talk between 2D images and 3D point clouds is managed using a cosine similarity loss. 

Another line of research has explored feature fusion, aiming to integrate geometric and radiometric features directly, before applying them to different downstream tasks. An early example is 3DMV~\cite{dai20183dmv}, which integrates image features from multi-view RGB images and geometric features for 3D semantic segmentation. 
ImVoteNet~\cite{qi2020imvotenet} introduces a joint 2D-3D voting scheme for 3D object detection. It first extracts features from images, projects these features back into 3D space, and concatenate them with point cloud features. Building on the geometry-only VoteNet~\cite{qi2019deep}, it then generates 3D Hough votes using the 3D points and the concatenated features to detect 3D objects.
In contrast, Pri3D~\cite{hou2021pri3d} proposes utilizing 3D priors for 2D image understanding. It demonstrates how 3D geometric pre-training can benefit complex 2D perception tasks such as semantic segmentation and object detection. 
Inspired by it, PCR-CG~\cite{zhang2022pcr} concatenates 3D point clouds with pixel-wise image features and passes them into a 3D backbone for training. The image features are extracted from a pre-trained model, without further retraining.
PointCMT~\cite{yan2022let} leverages radiometric features from 2D images to enhance 3D point cloud shape analysis. Meanwhile, DeepViewAgg~\cite{robert2022dva} introduces a view condition-based attention module to fuse multi-view features. It assigns per-pixel features to per-point features before passing them into a 3D backbone.

Different to existing cross-modal feature learning methods, the method we propose is a two-stage coarse-to-fine feature fusion approach \change{that leverages both pixel-wise image features (global type) and patch-wise image features (local type), specifically customized for 3D point cloud registration}. It consists of (i) assigning pixel-wise image features to 3D input point clouds to enhance learned 3D point features and (ii) fusing patch-wise image features with superpoint features for improving the coarse matching quality.
\section{Methodology}
\label{sec:method}

\begin{figure*}[ht]
  \centering
  \includegraphics[width=0.8\linewidth]{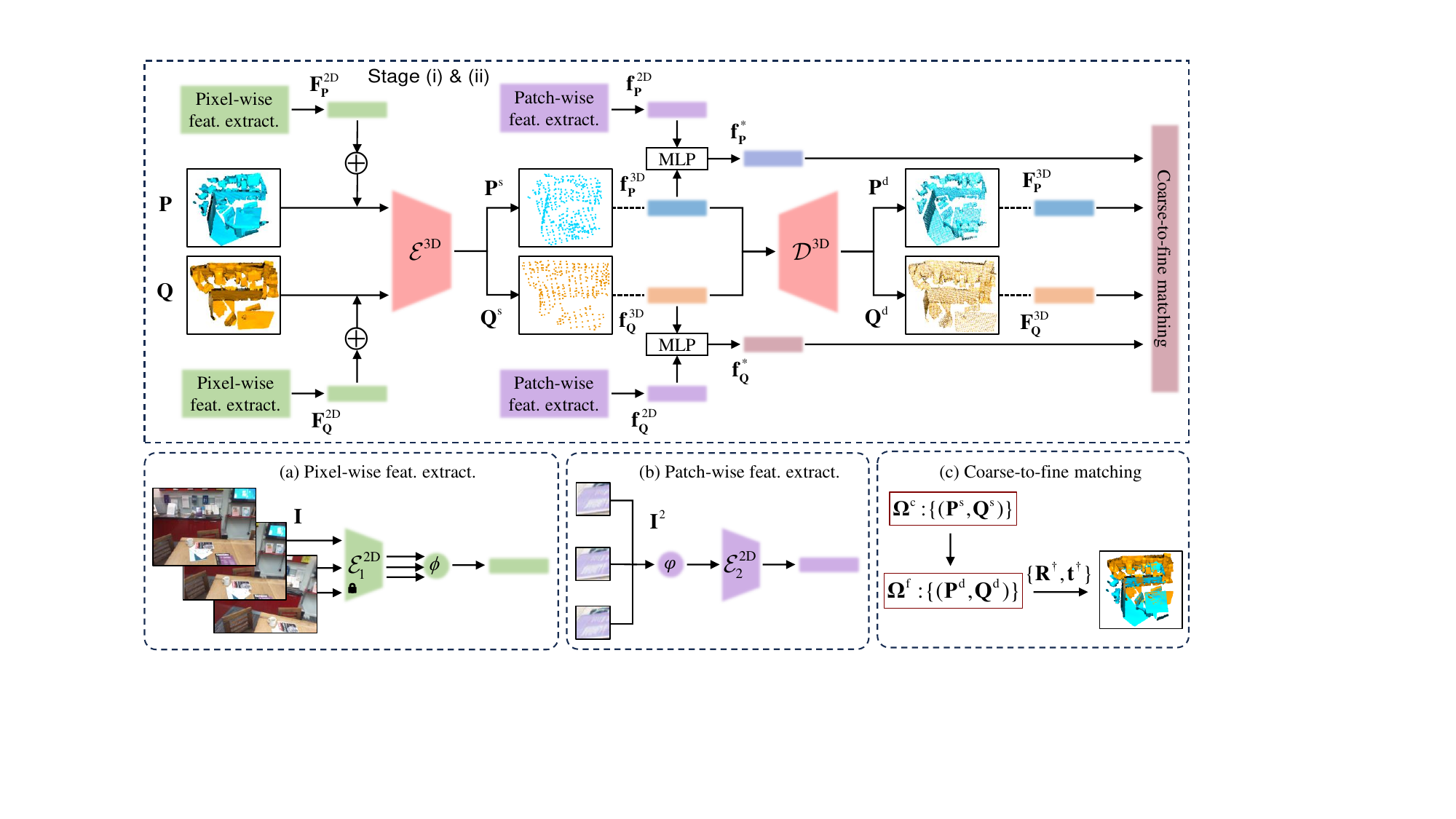}
  \caption{Overview of our method. Utilizing the KPConv-FPN backbone for point feature extraction, our method implements a two-stage, cross-modal feature fusion process. Stage (i) involves assigning pixel-wise image features $\mathbf{F}^{\mathrm{2D}}_{\mathbf{P}}$ to the 3D input point cloud $\mathbf{P}$. These are then processed through a 3D encoder $\mathcal{E} ^{\mathrm{3D}}$, yielding superpoints $\mathbf{P}^{\mathrm{s}}$ and their associated features $\mathbf{f}^{\mathrm{3D}}_{\mathbf{P}}$. In Stage (ii), the $\mathbf{f}^{\mathrm{3D}}_{\mathbf{P}}$ are integrated with patch-wise image features $\mathbf{f}^{\mathrm{2D}}_{\mathbf{P}}$, creating an enriched set of fused superpoint features $\mathbf{f}^{\mathrm{*}}_{\mathbf{P}}$ for coarse matching. Correspondences are established following the coarse-to-fine matching mechanism.
  \change{Module (a): Multiple pixel-wise features are derived using a frozen 2D pixel-wise model $\mathcal{E}^\mathrm{2D}_1$, and one of multiple pixel-wise features is then selected through a selection operation $\phi$.
  Module (b): Multiple patch-wise features are obtained using a 2D patch-wise model $\mathcal{E}^\mathrm{2D}_2$, and one of them is selected through a selection operation $\varphi$ fused with corresponding superpoint features $\mathbf{f}^{\mathrm{3D}}_{\mathbf{P}}$. 
  Module (c): Coarse correspondences $\mathbf{\Omega}^\mathrm{c}$ are initially established based on hybrid features extracted from geometric structure embedding and fused features $\mathbf{f}^\mathrm{*}_{\mathbf{P}}$ and $\mathbf{f}^\mathrm{*}_{\mathbf{Q}}$. These correspondences are then propagated to fine correspondences $\mathbf{\Omega}^\mathrm{f}$. Correspondences are used to estimate the transformation $\{\mathbf{R^{\dag}},\mathbf{t^{\dag}}\}$ using an LGR~\cite{qin2022geometric} estimator. 
  }}
  \label{fig:overview_of_pipeline}
\end{figure*}

\paragraph{\textnormal{\textbf{Problem statement}}} 

We are provided with two partially overlapping point clouds $\mathbf{P} = \{\mathbf{p}_i \in \mathbb{R}^3 \mid i = 1, ..., M\}$ and $\mathbf{Q} = \{\mathbf{q}_i \in \mathbb{R}^3 \mid i = 1, ..., N\}$, 
consisting of $M$ and $N$ points, respectively. $\mathbf{P}$ and $\mathbf{Q}$ are associated with a set of images $\mathbf{I}_\mathbf{P} = \{\mathbf{I}_{\mathbf{P},k} \in \mathbb{R}^{H\times W \times 3} \mid k = 1, ..., L_1\}$ and $\mathbf{I}_\mathbf{Q} = \{\mathbf{I}_{\mathbf{Q},k} \in \mathbb{R}^{H\times W \times 3} \mid k = 1, ..., L_2\}$, respectively, with known camera intrinsic and extrinsic parameters. Here, $H$ and $W$ are the image dimensions, respectively, while $L_1$ and $L_2$ denote the numbers of images available. The goal of point cloud registration is to estimate a rigid transformation $\mathbf{T}= \{\mathbf{R^\ast}, \mathbf{t^\ast} \}$ that aligns point cloud $\mathbf{P}$ with point cloud $\mathbf{Q}$ by solving the following optimization problem:

\begin{multline}
  (\mathbf{R^{\ast}},\mathbf{t^{\ast}}) = 
  \underset{(\mathbf{R},\mathbf{t}) \in SO(3) \times \mathbb{R}^3}{\mathrm{argmin}} \\
  \sum\nolimits_{(\mathbf{p}_{m_{j}},\mathbf{q}_{n_{j}}) \in \overline{\mathbf{\Omega}}} 
  {\left \| \mathbf{R} \cdot \mathbf{p}_{m_{j}} + \mathbf{t} - \mathbf{q}_{n_{j}} \right \|}^2_2,
  \label{eq:transform}
\end{multline}
where $\overline{\mathbf{\Omega}} = \{(\mathbf{p}_{m_{j}},\mathbf{q}_{n_{j}}) \mid j = 1, ..., O\}$ denotes the ground-truth correspondences.

Since $\overline{\mathbf{\Omega}}$ is typically unavailable in real-world scenarios, the task is often solved by learning feature descriptors and then establishing correspondences through feature matching. In this work, we use both point clouds and RGB images to learn feature descriptors.
\change{These RGB images can be captured using RGB-D sensors, standalone RGB cameras, or built-in calibrated RGB cameras in most terrestrial laser scanners. In all cases, we assume that the co-registration between images and point clouds, \ie, the estimation of camera intrinsic and extrinsic parameters, has been completed during preprocessing.} For instance, in terrestrial laser scanners with built-in RGB cameras, these parameters are typically provided after preprocessing data with manufacturer-supplied software.
Given 3D point correspondences subsequently established by feature matching, the transformation can be estimated using, \eg, RANSAC~\cite{81ransac}, the SVD-based Kabsch algorithm~\cite{kabsch1976solution,stewart1993svd}, or LGR~\cite{qin2022geometric}. Herein, we use the LGR~\cite{qin2022geometric} estimator.

\paragraph{\textnormal{\textbf{Overview}}} 

Our proposed method is illustrated in \cref{fig:overview_of_pipeline}. Below, we first describe the  3D point feature extraction (\S\ref{sec:pcd_part}), and the extraction of 2D image features (\S\ref{sec:img_part}). Then, we illustrate our cross-modal feature fusion (\S\ref{sec:fusion_part}). Finally, we describe
the coarse-to-fine feature matching and transformation estimation (\S\ref{sec:matching_part}). 


\subsection{3D point feature extraction}
\label{sec:pcd_part}

Following \cite{qin2022geometric}, we adopt the KPConv-FPN~\cite{Lin2016FPN,thomas2019KPConv} for the point feature extraction. It first employs hierarchical grid subsampling~\cite{thomas2019KPConv} to downsample the point cloud. The encoder $\mathcal{E} ^{\mathrm{3D}}$ takes the point cloud $\mathbf{P}$ and the pixel-wise image features $\mathbf{F}^{\mathrm{2D}}_{\mathbf{P}}$ (\cf \cref{sec:img_part}) as inputs. It then outputs superpoints $\mathbf{P}^{\mathrm{s}}$ and their associated features $\mathbf{f}^{\mathrm{3D}}_{\mathbf{P}}$ as depicted in \cref{eq:superpoint_feat}. 
The decoder $\mathcal{D} ^{\mathrm{3D}}$ combines nearest upsampling with linear layers. It also includes skip connections to pass the features between intermediate layers of the decoder and the corresponding encoder. The decoder outputs the dense points $\mathbf{P}^{\mathrm{d}}$ and their associated features $\mathbf{F}^{\mathrm{3D}}_{\mathbf{P}}$, as shown in \cref{eq:point_feat}.
\change{Both $\mathbf{P}^{\mathrm{s}}$ and $\mathbf{P}^{\mathrm{d}}$, along with their associated features are used in the matching process.}

\begin{equation}
  (\mathbf{P}^{\mathrm{s}}, \mathbf{f}^{\mathrm{3D}}_{\mathbf{P}}) = \mathcal{E} ^{\mathrm{3D}}(\mathbf{P}, \mathbf{F}^{\mathrm{2D}}_{\mathbf{P}}).
  \label{eq:superpoint_feat}
\end{equation}

\begin{equation}
  (\mathbf{P}^{\mathrm{d}}, {\mathbf{F}}^{\mathrm{3D}}_{\mathbf{P}}) = \mathcal{D}^{\mathrm{3D}}(\mathbf{P}^{\mathrm{s}}, \mathbf{f}^{\mathrm{3D}}_{\mathbf{P}}).
  \label{eq:point_feat}
\end{equation}

We obtain the superpoints and associated features for point cloud $\mathbf{Q}$ in the same way. To condense the notation, any definition or equation involving $\mathbf{P}, \mathbf{p}$ is implicitly understood to extend to an analogous version holding for $\mathbf{Q}, \mathbf{q}$ throughout the remainder of the paper.


\subsection{2D image feature extraction}
\label{sec:img_part}

\paragraph{\textnormal{\textbf{Projection from point cloud to image}}}

We consider a multi-view case in which each point cloud is associated with multiple RGB images. Given a 3D point $\mathbf{p}_i \in \mathbf{P}$, along with camera intrinsic and extrinsic matrices, $\mathbf{C}_{\mathrm{int},k}$ and $\mathbf{C}_{\mathrm{ext},k}$, which link $\mathbf{P}$ and image $\mathbf{I}_{\mathbf{P},k}$, we can project each $\mathbf{p}_i$ onto the image plane using homogeneous coordinates ~\cite{camera_model} to obtain the corresponding image coordinates $\mathbf{u}_i$, as shown in \cref{eq:projection}. After establishing the projection, we proceed to extract both pixel-wise and patch-wise image features from the RGB images.

\begin{equation}
    \mathbf{u}_i = \mathbf{C}_{\mathrm{ext},k} \cdot \mathbf{C}_{\mathrm{ext},k} \cdot \mathbf{p}_i,
    \label{eq:projection}
\end{equation}
where $\mathbf{u}_i = [u_i, v_i, 1]^\mathrm{T}$ and $\mathbf{p}_i = [X_i, Y_i, Z_i, 1]^\mathrm{T}$.

\paragraph{\textnormal{\textbf{Pixel-wise image feature extractor}}}

Directly concatenating low-level image features, \eg, RGB values, with point clouds often yields no significant or even negative benefits, as observed in previous studies~\cite{dai20183dmv,FCGF2019,qi2020imvotenet,zhang2022pcr}. This lack of benefit is likely due to the sensitivity of low-level image features to environmental conditions, \eg, illumination changes~\cite{d2net}. Therefore, similar to \cite{zhang2022pcr}, we use a 2D fully convolutional network $\mathcal{E}^\mathrm{2D}_1$ (a frozen pre-trained ResNet-50 \cite{resnet,hou2021pri3d}) to extract higher-dimensional pixel-wise image features \change{(see \cref{fig:overview_of_pipeline} (a))}. When multi-view images are available, especially in RGB-D datasets, multiple pixel-wise image features from different images are aggregated into $\mathbf{F}^{\mathrm{2D}}_{\mathbf{P}}$, as described in the paragraph on multi-view feature selection below.

\paragraph{\textnormal{\textbf{Patch-wise image feature extractor}}} 

We mitigate the problem caused by ambiguous geometry by leveraging patch-wise image features. Utilizing the 3D-2D projection, we map each superpoint and its 
neighborhood~\cite{hermosilla2018mccnn,thomas2019KPConv} (defined by a certain Euclidean radius)
onto multiple RGB images. We then extract multiple image patches $\mathbf{I}^{2}_\mathbf{P} = \{\mathbf{I}^2_{\mathbf{P},k} \in \mathbb{R}^{H_2\times W_2 \times 3} \mid k = 1, ..., L_1\}$ based on these projected pixels. One patch is selected as described in the subsequent paragraph and then resized and passed into the 2D patch-based network $\mathcal{E}^{\mathrm{2D}}_2$~\cite{pham2020lcd} to extract patch-wise image features $\mathbf{f}^{\mathrm{2D}}_{\mathbf{P}}$ \change{(see \cref{fig:overview_of_pipeline} (b))}. 

\paragraph{\textnormal{\textbf{Multi-view feature selection}}} 

\change{A point cloud may be associated with multiple RGB images captured from different viewpoints, and these images often overlap. In such cases, inconsistencies can arise when multiple image features extracted from different viewpoints are assigned to the same point.}
We address this by reducing multiple pixel-wise image features to a single one using a selection operation $\phi$. This operation ranks the images based on the spatial proximity between the 3D coordinates of the point cloud origin and the camera positions. The impact of different multi-view feature selection strategies is discussed in Section \ref{subsec:multiview_feature}. For multiple patches constructed from different images during patch-wise image feature extraction, the patch with the largest size is selected; we denote this selection operation as $\varphi$. 


\subsection{Cross-modal feature fusion}
\label{sec:fusion_part}

Given the input point cloud $\mathbf{P}$ along with superpoint features $\mathbf{f}^\mathrm{3D}_{\mathbf{P}}$, pixel-wise image features $\mathbf{F}^{\mathrm{2D}}_{\mathbf{P}}$, and patch-wise image features $\mathbf{f}^\mathrm{2D}_{\mathbf{P}}$, our proposed feature fusion module comprises two stages \change{(see \cref{fig:overview_of_pipeline}, top)}: (1) assigning $\mathbf{F}^{\mathrm{2D}}_{\mathbf{P}}$ to the 3D input point cloud $\mathbf{P}$, and 
(2) incorporating $\mathbf{f}^\mathrm{2D}_{\mathbf{P}}$ with $\mathbf{f}^\mathrm{3D}_{\mathbf{P}}$ using a simple MLP layer into the fused superpoint features  $\mathbf{f}^\mathrm{*}_{\mathbf{P}}$ via $\mathbf{f}^\mathrm{*}_{\mathbf{P}} = \mathrm{MLP}([\mathbf{f}^\mathrm{3D}_{\mathbf{P}}, \mathbf{f}^{\mathrm{2D}}_{\mathbf{P}}])$.


\subsection{Coarse-to-fine matching}
\label{sec:matching_part}

\change{For clarity, we implement the coarse-to-fine matching module (see \cref{fig:overview_of_pipeline} (c)) following \cite{qin2022geometric}, with two key differences regarding input features. First, the superpoint features $\mathbf{f}^\mathrm{3D}_{\mathbf{P}}$ from the 3D encoder $\mathcal{E} ^{\mathrm{3D}}$ are already enhanced by incorporating pixel-wise image features $\mathbf{F}^\mathrm{2D}_{\mathbf{P}}$ as input. Second, we use the fused superpoint features $\mathbf{f}^\mathrm{*}_{\mathbf{P}}$ instead of $\mathbf{f}^\mathrm{3D}_{\mathbf{P}}$ for coarse matching. Since this module is not the main focus of our contribution, we provide only the necessary details below, while further specifics are available in~\cite{qin2022geometric}.}

\paragraph{\textnormal{\textbf{Coarse matching}}}

Before fusing superpoint features $\mathbf{f}^\mathrm{3D}_{\mathbf{P}}$, $\mathbf{f}^\mathrm{3D}_{\mathbf{Q}}$ with $\mathbf{f}^\mathrm{2D}_{\mathbf{P}}$, $\mathbf{f}^\mathrm{2D}_{\mathbf{Q}}$, we follow~\cite{qin2022geometric,yu2023rotation} to refine $\mathbf{f}^\mathrm{3D}_{\mathbf{P}}$ and $\mathbf{f}^\mathrm{3D}_{\mathbf{Q}}$ using attention modules and normalize them onto the unit sphere. Next, we measure the similarity using a Gaussian correlation matrix $\mathbf{S}$, with $\mathbf{S}(i,j)=\mathrm{exp}(-\| \mathbf{f}^\mathrm{*}_{\mathbf{P},i} - \mathbf{f}^\mathrm{*}_{\mathbf{Q},j}\|^2_2)$. After a dual-normalization operation~\cite{Rocco18neigh,sun2021loftr,qin2022geometric} on $\mathbf{S}$, 
the largest $N_c$ entries in $\mathbf{S}$
are selected as a set of coarse correspondences $\mathbf{\Omega}^\mathrm{c}$. 

\paragraph{\textnormal{\textbf{Fine matching}}}

Once $\mathbf{\Omega}^\mathrm{c}$ is 
established, dense points $\mathbf{P}^{\mathrm{d}}$ and associated ${\mathbf{F}}^{\mathrm{3D}}_{\mathbf{P}}$ are assigned to their nearest superpoints in 3D space to construct superpoint patches $\mathbf{G}^{\mathbf{P}} = \{ \mathbf{G}^{\mathbf{P}}_i \in \mathbb{R}^{m \times 3} \mid i = 1, ..., M^{\mathrm{s}} \}$ and feature matrices $\mathbf{F}^{\mathbf{G}}_\mathbf{P} = \{ \mathbf{F}^{\mathbf{G}}_{\mathbf{P},i} \in \mathbb{R}^{m \times c} \mid i = 1, ..., M^{\mathrm{s}} \}$, with $M^\mathrm{s}$ the number of superpoints, $m$ the number of dense points lying in the current patch, and $c$ the dimension of dense point features. A point-to-node grouping strategy~\cite{sonet2018,yu2021cofinet} is utilized for the assignment. 
Specifically, a similarity matrix between $\mathbf{F}^{\mathbf{G}}_{\mathbf{P},i}$ and $\mathbf{F}^{\mathbf{G}}_{\mathbf{Q},j}$ is first calculated as $\mathbf{S}_l=\mathbf{F}^{\mathbf{G}}_{\mathbf{P},i}(\mathbf{F}^{\mathbf{G}}_{\mathbf{Q},j})^T/\sqrt{c}$ upon which we follow \cite{sarlin20superglue} to augment it for matching, run the Sinkhorn algorithm~\cite{Sinkhorn1967ConcerningNM} and extract fine correspondences $\mathbf{\Omega}^\mathrm{f}_i$ via mutual top-$k$ selection. Each $\mathbf{\Omega}^\mathrm{f}_i$ consists of matches of dense points located in the matched superpoint patches, and all of them are collected to form the final correspondences $\mathbf{\Omega}^\mathrm{f} = \cup^{N^{c}}_{i=1} \mathbf{\Omega}^\mathrm{f}_i$, with $N^{c}$ the number of superpoint matches.

After the coarse-to-fine matching, we apply the LGR~\cite{qin2022geometric} estimator to estimate the final transformation. Specifically, we first estimate each individual transformation $(\mathbf{R}_i,\mathbf{t}_i)$ from $\mathbf{\Omega}^\mathrm{f}_i$ following~\cref{eq:transform}. Among all the estimated transformations, we then select the one, denoted as $(\mathbf{R^{\dag}},\mathbf{t^{\dag}})$, with the most inlier correspondences according to \cref{eq:final_estimate}. Finally, based on all inlier correspondences, $(\mathbf{R^{\dag}},\mathbf{t^{\dag}})$ is iteratively refined using the SVD-based Kabsch algorithm~\cite{kabsch1976solution,stewart1993svd,92weightedsvd}.

\begin{multline}
  (\mathbf{R^{\dag}},\mathbf{t^{\dag}}) = 
  \underset{\mathbf{R}_i,\mathbf{t}_i}{\mathrm{argmax}} 
  \sum\nolimits_{(\mathbf{p}^{\mathrm{d}}_{m_{j}},\mathbf{q}^{\mathrm{d}}_{n_{j}}) \in {\mathbf{\Omega}}^{\mathrm{f}}_i} \\
  \big( \left \| \mathbf{R}_i \cdot \mathbf{p}^{\mathrm{d}}_{m_{j}} + \mathbf{t}_i - \mathbf{q}^{\mathrm{d}}_{n_{j}} \right \|^2_2 < \mathrm{\Delta} \big),
  \label{eq:final_estimate}
\end{multline}
where $\mathrm{\Delta}$ denotes the acceptance radius and $(\mathbf{R}_i, \mathbf{t}_i) \in SO(3)\times \mathbb{R}^3$.

\subsection{Training loss} 

To supervise superpoint correspondences in coarse matching, we follow \cite{qin2022geometric} and employ an overlap-aware circle loss, denoted as $\mathcal{L}_{\mathrm{point}}$. This loss is a variant of the more common triplet loss~\cite{FCGF2019}, which aims to minimize the distance between positively matched pairs of superpoint features. Similarly, we introduce the term $\mathcal{L}^{\mathrm{patch}}_{\mathrm{img}}$ to minimize the distance between positively matched pairs of patch-wise image features. The final dense point matching loss, denoted as $\mathcal{L}_{{\mathrm{\Omega}^{\mathrm{f}}}}$, is a negative log-likelihood loss~\cite{sarlin20superglue,qin2022geometric,yu2023rotation} computed over all sampled superpoint matches. We train our model by minimizing a combined loss function, as shown in~\cref{eq:loss_func}. Further details about these loss functions are provided in \ref{sec:sup_loss}.

\begin{equation}
  \mathcal{L} = \mathcal{L}_{\mathrm{point}} + \mathcal{L}^{\mathrm{patch}}_{\mathrm{img}} + \mathcal{L}_{{\mathrm{\Omega}^{\mathrm{f}}}},
  \label{eq:loss_func}
\end{equation}

\subsection{\change{Implementation details}}
\label{sec:sup_implement}

We train the network on 3DMatch~\cite{zeng20163dmatch} dataset using a single GeForce RTX 4090 GPU. The network is trained with the Adam~\cite{adam15} optimizer for 20 epochs on 3DMatch~\cite{zeng20163dmatch}. We set the batch size to 1 and implement a weight decay of $10^{-6}$. The initial learning rate is set at $10^{-4}$, with an exponential decay of 0.05 applied after each epoch for 3DMatch~\cite{zeng20163dmatch}. We adopt the same data augmentation as in~\cite{predator2021,qin2022geometric}. 
For pixel-wise image feature extraction, we employ a feature dimension of 128, and 3D points lacking corresponding image pixels are assigned a 128-dimensional vector filled with ones. In constructing image patches, patches are resized to $64 \times 64$, and the dimension of extracted patch-wise image features is 256. \change{Additionally, the superpoint features $\mathbf{f}^\mathrm{3D}_{\mathbf{P}}$, fused superpoint features $\mathbf{f}^\mathrm{*}_{\mathbf{P}}$, and dense point features $\mathbf{F}^\mathrm{3D}_{\mathbf{P}}$ all have a dimension of 256}. During training, we randomly select 100 pairs of ground-truth coarse correspondences for image patch construction. While during testing, we construct image patches for all pairs of estimated coarse correspondences. For other hyper-parameters, we use the same with GeoTransformer~\cite{qin2022geometric}. \change{We apply data augmentation during training, including adding Gaussian noise with a standard deviation of 0.005, random scaling with uniform distribution [0.9, 1.1], and random rotation around an arbitrary axis by an angle drawn from $[0^\circ, 360^\circ)$~\cite{bai2020d3feat}~\footnote{The model used for extracting pixel-wise image features was trained with color space augmentation including RGB jittering, random color dropping, and Gaussian blur~\cite{hou2021pri3d}, while the model for extracting patch-wise image features was trained under varying lighting conditions~\cite{pham2020lcd}.}.}

\section{Experiments and results}
\label{sec:experi}

In this section, we conduct extensive experiments on four open-source datasets to evaluate the proposed method. First, we describe details about datasets we used in Section \ref{sec:dataset}. Then, we illustrate the evaluation setup in Section \ref{sec:setup}. Next, we compare with other registration methods on common datasets in Section \ref{subsec:main_results}. Finally, we evaluate our method on our subset datasets that feature the geometrically ambiguous challenge in Section \ref{sec:subset_results}.

\subsection{Dataset description}
\label{sec:dataset}

We evaluate our method mainly on four open-source datasets: 3DMatch~\cite{zeng20163dmatch}, 3DLoMatch~\cite{predator2021}, 
IndoorLRS~\cite{Park2017coloricp} and ScanNet++~\cite{yeshwanthliu2023scannetpp}. 
We train the model only on 3DMatch dataset, and directly evaluate it on other datasets. An overview of these datasets is given in \cref{tab:dataset_overview}. \change{These four datasets are well-established and easily accessible benchmarks, making them useful for the present research. But they do not capture the complexities of more challenging outdoor cases. We leave further investigation into outdoor scenes to future research and provide encouraging results from applying our method to a publicly accessible outdoor dataset in~\ref{sec:outdoor}.}

\begin{table*}[h]
    \caption{Overview of the datasets we used.}
    \vspace{0.5em}
  \label{tab:dataset_overview}
  \centering
  \resizebox{0.8\textwidth}{!}{
  \begin{tabular}{lcccccc}
   \toprule
    \multirow{2}{*}{Dataset} & \multirow{2}{*}{Sensor} & \multirow{2}{*}{RGB resolution (pixels)} & \multicolumn{4}{c}{Number of \change{point cloud} pairs} \\
    & & & Training & Validation & Test & Subset \\
    \midrule
    3DMatch~&~Kinect RGB-D sensor~&~480 x 640~&~20,642~&~1,331~&~1,623~&~161~\\
    3DLoMatch~&~Kinect RGB-D sensor~&~480 x 640~&~-~&-~&~1,781~&~534~\\
    IndoorLRS~&~Asus Xtion Live RGB-D sensor~&~480 x 640~&~-~&~-~&~8,525~&~572~\\
    ScanNet++~&~iPhone 13 Pro~&~1440 x 1920~&~-~&~-~&~19,284~&~2,784~\\
  \bottomrule
  \end{tabular}
  }
\end{table*}

\paragraph{\textnormal{\textbf{3DMatch and 3DLoMatch}}}
The 3DMatch~\cite{zeng20163dmatch} dataset is commonly used for the evaluation of the point cloud registration task, with data captured by the Kinect RGB-D sensor. The full dataset contains 62 scenes which are split 
into training, validation, and test sets with 46, 8, and 8 scenes, respectively. We use the training and validation sets preprocessed by~\cite{predator2021} and evaluate on both 3DMatch~\cite{zeng20163dmatch} and 3DLoMatch~\cite{predator2021} test sets. 
\change{A point cloud fragment refers to a fused point cloud generated from multiple consecutive RGB-D frames, typically integrated using truncated signed distance function (TSDF) volumetric fusion~\cite{predator2021}. For the 3DMatch and 3DLoMatch datasets, each point cloud fragment is reconstructed from 50 ordered RGB-D frames, following the approach used in previous works~\cite{predator2021,yu2021cofinet,qin2022geometric,zhang2022pcr} evaluated on the 3DMatch dataset. The registration evaluation is conducted on different pairs of these fragments. For consistency, we refer to ``point cloud fragments" as ``point clouds" and ``point cloud fragment pairs" as ``point cloud pairs" unless further specification is necessary.} 
The overlap ratio of point cloud pairs in 3DMatch~\cite{zeng20163dmatch} exceeds $30\%$, whereas in 3DLoMatch~\cite{predator2021}, it lies between $10\%$ and $30\%$. For each point cloud, the RGB images related to the first, middle and last frames are used to extract image features. The same applies to the IndoorLRS~\cite{Park2017coloricp} and ScanNet++~\cite{yeshwanthliu2023scannetpp} datasets. 

\paragraph{\textnormal{\textbf{IndoorLRS}}}


The Indoor LiDAR-RGBD Scan~\cite{Park2017coloricp} (IndoorLRS) dataset is a dataset with rich color texture. We use the RGB-D data captured by an Asus Xtion Live RGB-D sensor, with point clouds fused using 100 ordered RGB-D frames~\cite{Park2017coloricp}. We use all five scenes for testing, resulting in 8,525 \change{point cloud} pairs, each with an overlap ratio of more than 10\%. 

\paragraph{\textnormal{\textbf{ScanNet++}}}

\begin{figure}[h]
  \centering
  \includegraphics[width=1.0\linewidth]{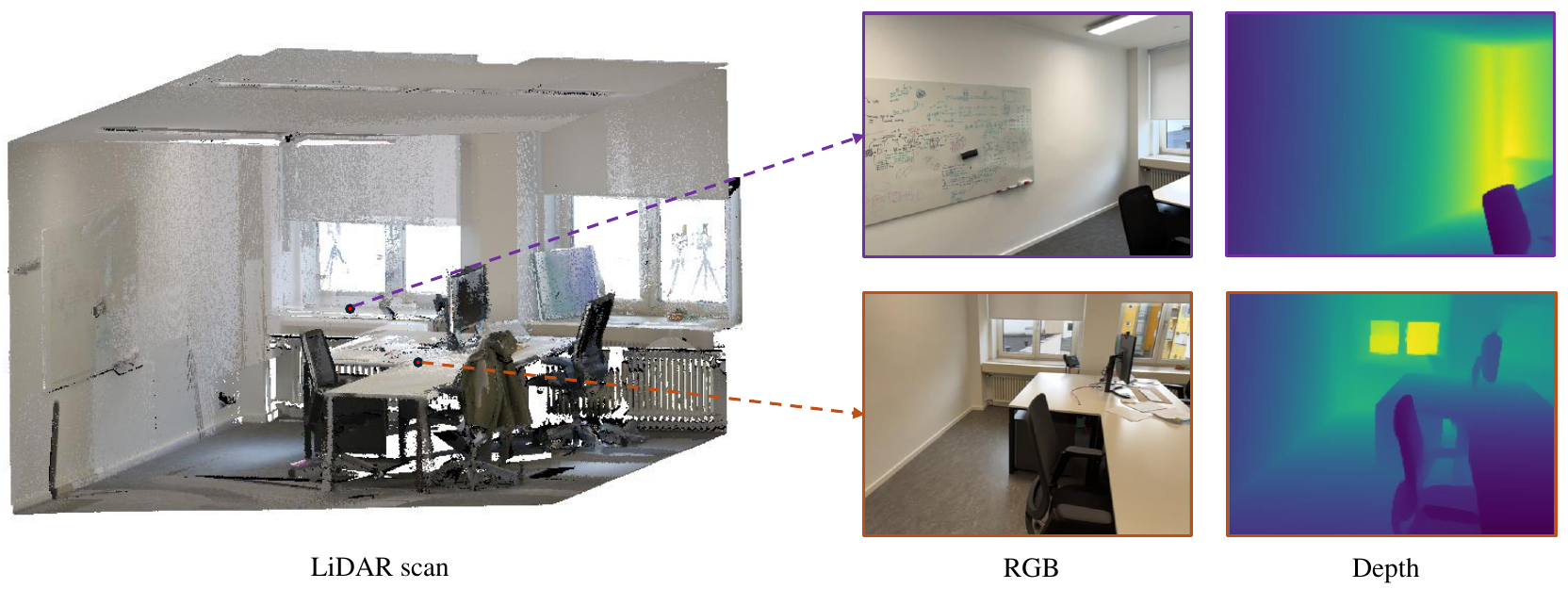}
  \caption{Dataset visualization (e.g., ScanNet++). Left: a scene scan capture by a LiDAR scanner; Right: two RGB and depth frames captured by a Iphone 13 Pro from two different locations. RGB frames exist large illumination changes, making it more challenging to use them.}
  \label{fig:scannet_situation}
\end{figure}

The ScanNet++~\cite{yeshwanthliu2023scannetpp} dataset is a recently released challenging dataset and primarily used for scene reconstruction and understanding (\cf \cref{fig:scannet_situation}). We preprocess the dataset and customize it for the point cloud registration task. We use the RGB-D frames captured by a commodity-level iPhone 13 Pro. These RGB-D frames are first processed to generate point cloud frames using provided camera intrinsic parameters. Every 20 point cloud frames are then integrated to one point cloud based on TSDF volumetric fusion using provided frame extrinsic parameters. We use the test set including 50 scenes for testing. For each scene, we select the first 50 point clouds to generate point cloud pairs with an overlap ratio of more than 10\%. Finally, we generate 19,284 point cloud pairs on ScanNet++~\cite{yeshwanthliu2023scannetpp} for the point cloud registration task.

\paragraph{\textnormal{\textbf{Subset datasets}}}


To assess the capability of our method in handling scenarios when geometry is ambiguous (\eg, planar), we select subsets from the aforementioned datasets. For each pair of point clouds aligned using the ground-truth transformation, we identify an overlapping region $\mathbf{O}$ using the nearest neighbor method. We then apply RANSAC for single-plane fitting. Each point $\mathbf{o}_i$ in $\mathbf{O}$ is checked against a distance threshold $\tau_1$ from the fitted plane $\mathbf{S}$. Points within $\tau_1$ are considered inliers, counted as $\mathrm{Z}^{o}$ in \cref{eq:plan_fitting}.

\begin{equation}
  \mathrm{Z}^{o} = {\sum^{\mathrm{Z}}\limits_{i=1}{(\left \| \mathbf{o}_i - \mathbf{S} \right \|}_2 < \mathrm{\tau_1})},
  \label{eq:plan_fitting}
\end{equation}
where $\mathrm{Z}$ is the number of points in the overlap.

We calculated a score $r=\mathrm{Z}^{o}/\mathrm{Z}$ for each pair to determine the proportion of inliers to total points in $\mathbf{O}$. Pairs with $r$ above a threshold $\tau_2$ are deemed geometrically planar.
For our evaluations, we determine the threshold $\tau_2$ depending on the geometric ambiguity situation of the corresponding datasets. Specifically, we extract subsets from the 3DMatch~\cite{zeng20163dmatch}, 3DLoMatch~\cite{predator2021} and ScanNet++~\cite{yeshwanthliu2023scannetpp} datasets by selecting point cloud pairs with $r$ above $\tau_2=0.7$, resulting in 161, 534 and 2,784 pairs, respectively. Additionally, we create the IndoorLRS\_Planar subset by applying $\tau_2=0.8$, which yields a collection of 572 point cloud pairs.


\subsection{Evaluation setup}
\label{sec:setup}

\subsubsection{Evaluation metrics}
\label{sec:metric}

We use \emph{Registration Recall} (RR) as the main metric across all four datasets, as it reflects the actual objective of point cloud registration. Following \cite{FCGF2019,predator2021,qin2022geometric}, we also report \emph{Feature Match Recall} (FMR), \emph{Inlier Ratio} (IR), \emph{Rotation Error} (RE) and \emph{Translation Error} (TE) for evaluation on all four datasets. \change{To ensure a fair comparison, we use the same thresholds to compute these metrics as in other baselines. Further analysis on the performance of different methods over a wide range for the allowable inlier distances, minimum inlier ratios, and minimum RMSEs can be found in~\ref{sec:sup_differ_thresholds}.}

\paragraph{\textnormal{\textbf{Inlier Ratio (IR)}}}

IR denotes the fraction of inlier correspondences among all estimated dense point correspondences $\mathbf{\Omega}^{\mathrm{f}}$.
A correspondence is considered as an inlier if the distance between two matched 3D points is smaller than a threshold $\mathrm{\Delta}_r=0.10$ m under the ground-truth transformation $(\mathbf{R^{\ast}},\mathbf{t^{\ast}})$:

\begin{equation}
  \mathrm{IR} = \frac{1}{|\mathbf{\Omega}^\mathrm{f}|}
  {\sum\nolimits_{(\mathbf{p}^{\mathrm{d}}_{m_{j}},\mathbf{q}^{\mathrm{d}}_{n_{j}}) \in {\mathbf{\Omega}^\mathrm{f}}}{\big (\left \| \mathbf{R}^{\ast} \cdot \mathbf{p}^{\mathrm{d}}_{m_{j}} + \mathbf{t}^{\ast} - \mathbf{q}^{\mathrm{d}}_{n_{j}} \right \|}_2 < \mathrm{\Delta}_r \big )}.
  \label{eq:inlier_ratio}
\end{equation}

\paragraph{\textnormal{\textbf{Feature Match Recall (FMR)}}} \change{FMR denotes the fraction of point cloud pairs for which the IR exceeds a predefined threshold $\tau_2 = 5\%$, as depicted in~\cref{eq:fmr}.} This metric measures the likelihood that an accurate transformation can be estimated with a robust estimator, \eg, RANSAC~\cite{81ransac}, the SVD-based Kabsch algorithm~\cite{kabsch1976solution,stewart1993svd}, or LGR~\cite{qin2022geometric}. However, FMR does not check if the transformation can actually be determined from those correspondences, as they may lie close together or along a straight edge~\cite{predator2021}. \change{To address this limitation, we introduce additional metrics in the following paragraphs.}

\begin{equation}
\mathrm{FMR} = \frac{1}{M} \sum_{i=1}^{M} (\mathrm{IR}_i > \tau_2),
\label{eq:fmr}
\end{equation}
where $M$ is the number of point cloud pairs.

\paragraph{\textnormal{\textbf{Registration recall (RR)}}} \change{Since we evaluate registration methods on thousands of point cloud pairs, reporting the average RMSE across all pairs may be misleading. This is because a single failed registration with a high RMSE could disproportionately affect the average, even if the method performs well on most pairs. Therefore, similar as other baselines, we compute the RR, which we consider} the most reliable metric as it reflects the actual task of point cloud registration. The RR for the four datasets used is defined as the fraction of point cloud pairs whose RMSE is below an acceptance radius 0.2 m. The RMSE is computed as.  

\begin{equation}
  \mathrm{RMSE} = \sqrt{\frac{1}{ |\overline{\mathbf{\Omega}}| }
  {\sum\nolimits_{(\mathbf{p}^{\mathrm{d}}_{m_{j}},\mathbf{q}^{\mathrm{d}}_{n_{j}}) \in {\overline{\mathbf{\Omega}}}}{\big (\left \| \mathbf{R}^{\dag} \cdot \mathbf{p}^{\mathrm{d}}_{m_{j}} + \mathbf{t}^{\dag} - \mathbf{q}^{\mathrm{d}}_{n_{j}} \right \|}^2_2 \big )}},
  \label{eq:rmse}
\end{equation}
where $\overline{\mathbf{\Omega}}$ denotes the ground-truth correspondences, and $(\mathbf{R^{\dag}},\mathbf{t^{\dag}})$ denotes the estimated transformation.

\paragraph{\textnormal{\textbf{Rotation Error (RE)}}}

RE represents the rotation deviation between estimated and ground-truth rotation matrices \change{(in $^\circ$)}:

\begin{equation}
  \mathrm{RE} = \arccos\left(\frac{\mathrm{trace}((\mathbf{R}^\dag)^\mathrm{T} \cdot \mathbf{R}^{\ast}) - 1}{2}\right).
  \label{eq:coarse_matching}
\end{equation}

\paragraph{\textnormal{\textbf{Translation Error (TE)}}}

TE represents the Euclidean distance between estimated and ground-truth translation vectors \change{(in $m$)}:

\begin{equation}
\mathrm{TE} = \lVert \mathbf{t}^{\dag} - \mathbf{t}^{\ast} \rVert_2.
\end{equation}

\subsubsection{Baselines} 

We compare CoFF to \change{different baselines, including two main ones:} PCR-CG~\cite{zhang2022pcr}, which uses both point clouds and RGB images, and GeoTransformer~\cite{qin2022geometric}, which utilizes only point cloud but is
the baseline we follow for our 3D point feature extraction. 
Both CoFiNet~\cite{yu2021cofinet} and GeoTransformer~\cite{qin2022geometric} employ a coarse-to-fine matching strategy similar to our method, \ie, \change{detector-free} methods.
For the 3DMatch~\cite{zeng20163dmatch} and 3DLoMatch~\cite{predator2021} datasets, we directly report the numbers from the respective publications. While for IndoorLRS~\cite{Park2017coloricp}, recently released ScanNet++~\cite{yeshwanthliu2023scannetpp}, and our specifically subset datasets, we use the released models (or the model we re-trained when it is not publicly available) for evaluation.
\change{All methods use 5,000 sample points for computing the IR and FMR on these datasets.} 


\subsection{Results on common datasets}
\label{subsec:main_results}


In \cref{tab:3dmatch} and \cref{tab:indoorlrs_scannetpp}, we present the registration results of our method CoFF and other baseline methods across four common datasets. As shown in these tables, our method consistently outperforms other methods in terms of FMR and RR across all datasets. 
Specifically, on the 3DMatch~\cite{zeng20163dmatch}  dataset, CoFF outperforms other methods by at least 1.4\% in FMR and 4.4\% in RR. Similarly, on the 3DLoMatch~\cite{predator2021} dataset, CoFF achieves improvements of 5.3\% in FMR and 7.6\% in RR over GeoTransformer, which is the second-best method. 
The IndoorLRS~\cite{Park2017coloricp} dataset presents a different scenario due to the use of point clouds fused from 100 frames, resulting in each point cloud containing richer features. Consequently, the performance gains of CoFF are relatively smaller on this dataset. Despite this, CoFF still achieves the highest FMR and surpasses CoFiNet~\cite{yu2021cofinet} by 0.4\% in RR.
On the ScanNet++~\cite{yeshwanthliu2023scannetpp} dataset, CoFF outperforms the second-best methods by 2.1\% in FMR and 5.3\% in RR. These results consistently demonstrate the effectiveness of our method across diverse and challenging datasets.

\begin{table}[h]
    \caption{Results on 3DMatch and 3DLoMatch datasets. We use \textBF{boldface} and \underline{underline} to highlight the best and second-best results in each column  (this holds for all tables, herein).}
    \vspace{0.5em}
  \label{tab:3dmatch}
  \centering
  \resizebox{0.99\columnwidth}{!}{
  \begin{tabular}{l|ccc|ccc}
    \toprule
     & \multicolumn{3}{c|}{3DMatch}&  \multicolumn{3}{c}{3DLoMatch}\\
    Method & FMR (\%)$\uparrow$ & IR (\%)$\uparrow$ & RR (\%)$\uparrow$ & FMR (\%)$\uparrow$ & IR (\%)$\uparrow$ & RR (\%)$\uparrow$ \\
    \midrule
    FCGF~\cite{FCGF2019}~&~97.4~&~56.8~&~85.1~~~&~~76.6~&~21.4~&~40.1~\\
    Predator~\cite{predator2021}~&~96.6~&~58.0~&~89.0~~~&~~78.6~&~26.7~&~59.8~\\
    CoFiNet~\cite{yu2021cofinet}~&~\underline{98.1}~&~49.8~&~89.3~~~&~~83.1~&~24.4~&~67.5~\\

    GeoTransformer~\cite{qin2022geometric}~&~97.9~&~\underline{71.9}~&~\underline{91.5}~~~&~~\underline{88.3}~&~\underline{43.5}~&~\underline{74.0}~\\
    PCR-CG~\cite{zhang2022pcr}~&~97.4~&~-~&~89.4~~~&~~80.4~&~-~&~66.3~\\
    PCR-CG~\cite{zhang2022pcr}(retrain)~&~97.2~&~57.4~&~\underline{91.5}~~~&~~77.7~&~25.6~&~63.8~\\
    CoFF(ours)~&~\textBF{99.5}~&~\textBF{74.3}~&~\textBF{95.9}~~~&~~\textBF{93.6}~&~\textBF{47.4}~&~\textBF{81.6}~\\
  \bottomrule
  \end{tabular}
  }
\end{table}

\begin{table}[h]
    \caption{Results on IndoorLRS and ScanNet++ datasets.}
    \vspace{0.5em}
  \label{tab:indoorlrs_scannetpp}
  \centering
  \resizebox{0.99\columnwidth}{!}{
  \begin{tabular}{l|ccc|ccc}
    \toprule
     & \multicolumn{3}{c|}{IndoorLRS}&  \multicolumn{3}{c}{ScanNet++}\\
    Method & FMR (\%)$\uparrow$ & IR (\%)$\uparrow$ & RR (\%)$\uparrow$ & FMR (\%)$\uparrow$ & IR (\%)$\uparrow$ & RR (\%)$\uparrow$ \\
    \midrule
    FCGF~\cite{FCGF2019}~&~\textBF{99.7}~&~62.6~&~89.6~~~&~~47.7~&~19.8~&~41.1~\\
    Predator~\cite{predator2021}~&~\underline{99.2}~&~60.5~&~95.4~~~&~~68.2~&~23.8~&~63.7~\\
    CoFiNet~\cite{yu2021cofinet}~&~\textBF{99.7}~&~51.0~&~\underline{96.7}~~~&~~80.6~&~31.4~&~71.1~\\
    GeoTransformer~\cite{qin2022geometric}~&~\textBF{99.7}~&~\textBF{74.1}~&~96.2~~~&~~81.2~&~\underline{37.2}~&~\underline{73.4}~\\    
    PCR-CG~\cite{zhang2022pcr}(retrain)~&~97.5~&~54.2~&~92.7~~~&~~\underline{86.2}~&~\textBF{41.7}~&~61.5~\\
    CoFF(ours)~&~\textBF{99.7}~&~\underline{71.7}~&~\textBF{97.1}~~~&~~\textBF{88.3}~&~34.1~&~\textBF{78.7}~\\
  \bottomrule
  \end{tabular}
  }   
\end{table}

We also observe that on the ScanNet++~\cite{yeshwanthliu2023scannetpp} dataset (\cf \cref{tab:indoorlrs_scannetpp}), our method has a lower IR compared to GeoTransformer. However, it still achieves the highest RR among the baseline methods.
This phenomenon \change{(also appears on other datasets, \eg, ScanNet++\_Planar in~\cref{tab:indoor_scannetpp_planar})}, where a high IR does not always correspond to a high RR, may occur because correspondences can cluster together, as also observed in previous works such as Predator~\cite{predator2021} and GeoTransformer~\cite{qin2022geometric}. \change{This issue may particularly appear in detector-based methods, when detected keypoints are not spatially evenly distributed across the scene.}

\begin{figure*}[tb]
  \centering
  \includegraphics[width=0.8\linewidth]{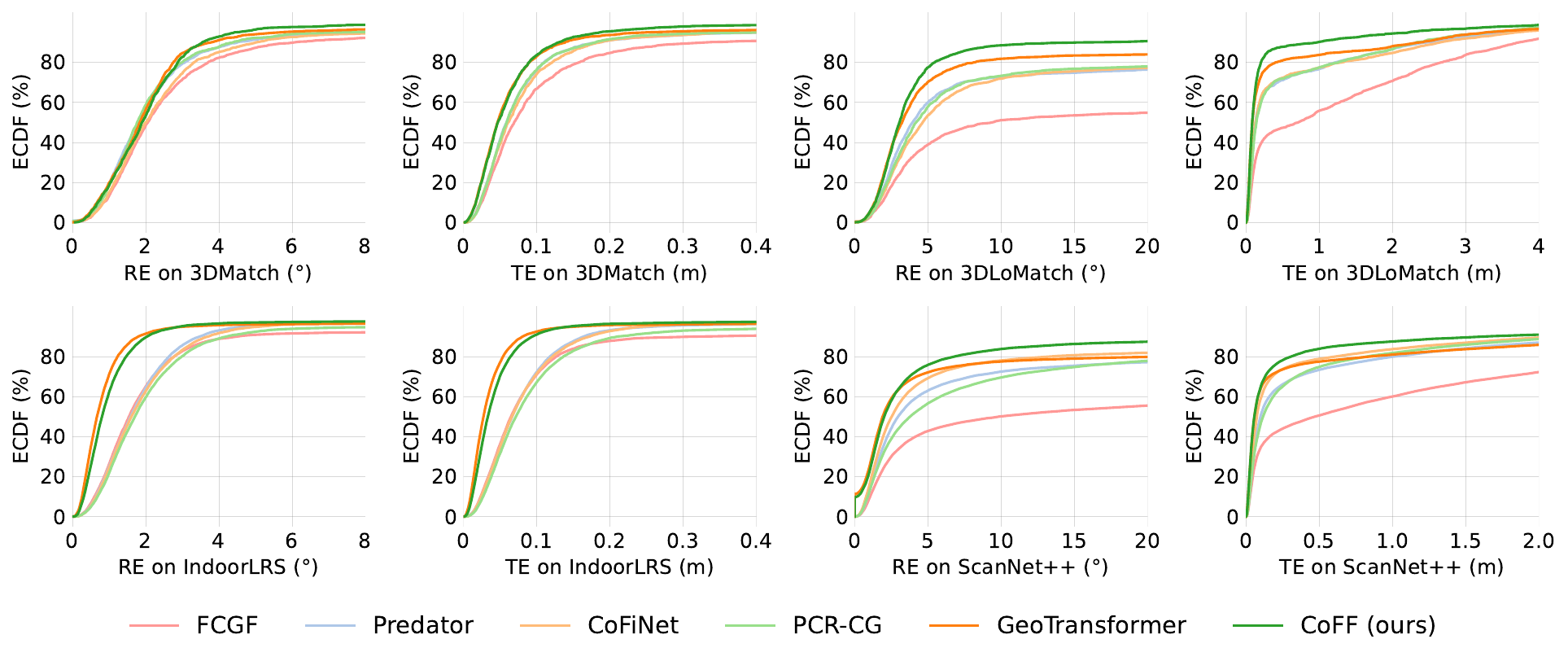}
  \caption{Rotation Error (RE) and Translation Error (TE) for all point cloud pairs on four common datasets. To enhance visualization, we apply different x-axis value ranges for each of the four datasets based on their specific characteristics.}
  \label{fig:re_te_common}
\end{figure*}

To better understand the error distribution, we present pair-wise registration results of \change{RE} and TE across all datasets. \change{However, using mean values of RE and TE across all point cloud pairs (e.g., 1,623 pairs in 3DMatch) may obscure the actual registration performance, as failed pairs with large RE and TE could inflate these metrics. To address this, we choose to provide a more comprehensive analysis by using Empirical Cumulative Distribution Function (ECDF) curves in~\cref{fig:re_te_common} and~\cref{fig:re_te_subset}, which report RE and TE for all individual pairs.} 
Our method demonstrates superior performance across all datasets, as indicated by higher ECDF values at given threshold values of either RE or TE. This shows that our method consistently achieves lower error rates compared to the baseline methods.



\subsection{Results on subset datasets}
\label{sec:subset_results}

\begin{table}[ht]
    \caption{Results on 3DMatch\_Planar and 3DLoMatch\_Planar subset datasets.}
    \vspace{0.5em}  
  \label{tab:3dmatch_planar}
  \centering
  \resizebox{0.9\columnwidth}{!}{
  \begin{tabular}{l|ccc|ccc}
    \toprule
     & \multicolumn{3}{c|}{3DMatch\_Planar}&  \multicolumn{3}{c}{3DLoMatch\_Planar}\\
    Method & FMR (\%)$\uparrow$ & IR (\%)$\uparrow$ & RR (\%)$\uparrow$ & FMR (\%)$\uparrow$ & IR (\%)$\uparrow$ & RR (\%)$\uparrow$ \\
    \midrule
    FCGF~\cite{FCGF2019}~&~84.5~&~39.1~&~45.5~~~&~~61.6~&~14.6~&~27.4~\\
    Predator~\cite{predator2021}~&~83.4~&~39.8~&~55.5~~~&~~68.2~&~19.1~&~47.9~\\
    CoFiNet~\cite{yu2021cofinet}~&~89.9~&~31.4~&~65.2~~~&~~73.2~&~17.4~&~52.6~\\
    GeoTransformer~\cite{qin2022geometric}~&~\underline{96.5}~&~\underline{57.1}~&~\underline{76.4}~~~&~~\underline{75.8}~&~\underline{32.7}~&~\underline{59.5}~\\
    PCR-CG(retrain)~\cite{zhang2022pcr}~&~80.7~&~38.0~&~64.1~~~&~~64.5~&~17.4~&~48.1~\\
    CoFF(ours)~&~\textBF{99.0}~&~\textBF{60.8}~&~\textBF{90.5}~~~&~~\textBF{87.2}~&~\textBF{36.5}~&~\textBF{70.4}~\\
  \bottomrule
  \end{tabular}
  }
\end{table}

\begin{table}[ht]
    \caption{Results on IndoorLRS\_Planar and ScanNet++\_Planar subset datasets.}
    \vspace{0.5em}
  \label{tab:indoor_scannetpp_planar}
  \centering
  \resizebox{0.99\columnwidth}{!}{
  \begin{tabular}{l|ccc|ccc}
    \toprule
     & \multicolumn{3}{c|}{IndoorLRS\_Planar}&  \multicolumn{3}{c}{ScanNet++\_Planar}\\
    Method & FMR (\%)$\uparrow$ & IR (\%)$\uparrow$ & RR (\%)$\uparrow$ & FMR (\%)$\uparrow$ & IR (\%)$\uparrow$ & RR (\%)$\uparrow$ \\
    \midrule
    FCGF~\cite{FCGF2019}~&~\underline{98.6}~&~52.8~&~60.8~~~&~35.9~&~9.9~&~15.1~\\
    Predator~\cite{predator2021}~&~96.0~&~48.9~&~76.9~~~&~37.8~&~9.6~&~27.2~\\
    CoFiNet~\cite{yu2021cofinet}~&~97.2~&~44.2~&~87.3~~~&~53.3~&~13.7~&~33.4~\\
    GeoTransformer~\cite{qin2022geometric}~&~\underline{98.6}~&~\textBF{64.5}~&~\underline{91.2}~~~&~49.9~&~14.6~&~\underline{37.6}~\\
    PCR-CG(retrain)~\cite{zhang2022pcr}~&~91.8~&~39.6~&~68.3~~~&~\underline{67.9}~&~\textBF{26.7}~&~31.9~\\
    CoFF(ours)~&~\textBF{99.2}~&~\underline{60.5}~&~\textBF{94.2}~~~&~\textBF{70.6}~&~\underline{20.0}~&~\textBF{56.0}~\\
    
  \bottomrule
  \end{tabular}
  }
\end{table}

To further address the challenge of geometric ambiguity in existing registration methods, we evaluate CoFF and other baseline methods on subset datasets derived from four common datasets, as shown in \cref{tab:3dmatch_planar} and \cref{tab:indoor_scannetpp_planar}. 
Our method outperforms the second-best method by 2.5\% and 14.1\% in FMR and RR, respectively, on 3DMatch\_Planar. On 3DLoMatch\_Planar, which features both low overlap and geometrically ambiguous challenges, our method exceeds other baselines by at least 11.4\% in FMR and 10.9\% in RR. 
Similar substantial improvements are observed on Indoor\_Planar and ScanNet++\_Planar in \cref{tab:indoor_scannetpp_planar}.
Compared to the results on the common datasets, our method outperforms the baselines by a larger margin in both FMR and RR on the subset datasets. This highlights the capability of CoFF to handle scenarios with geometrically challenging conditions more effectively than existing methods.

\begin{figure*}[h]
  \centering
  \includegraphics[width=0.8\linewidth]{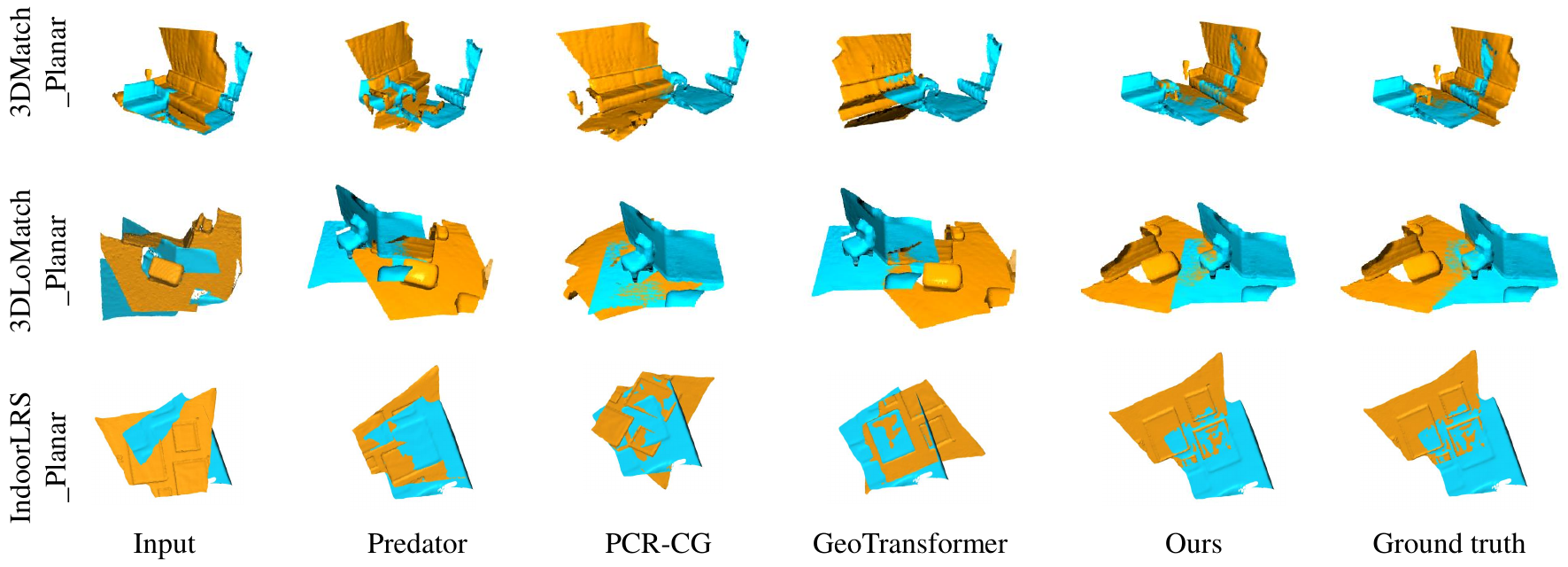}
  \caption{Qualitative results on different subset datasets. The first two rows depict point cloud pairs exhibiting symmetries, for which CoFF outperforms the other methods. The last row showcases point cloud pairs overlapping in the planar regions with CoFF the only method succeeding in these conditions. Despite high overlap ratios (60\% and 40\%) in the first and last rows, registration remains challenging for existing methods.
  }
  \label{fig:qualitative_results}
\end{figure*}

We also present some qualitative results of the subset datasets in \cref{fig:qualitative_results}, demonstrating the capability of our method to reliably align point clouds in challenging scenarios. Our observations reveal that ambiguous geometry can persist even with high overlap ratios, underscoring the importance of robust registration methods capable of handling such complexities.

\section{Discussion}
\label{sec:discuss}

In this discussion, we first analyze the factors contributing to the strong performance of CoFF in geometrically ambiguous cases in Section \ref{subsec:factors}. We then examine the effect of two different types of image features to the point cloud registration task in Section \ref{subsec:differ_img_feats}. Next, we explore the impact of different multi-view image feature selections in Section \ref{subsec:multiview_feature}. Further, we investigate the possibility of incorporating CoFF with different 3D registration methods in Section \ref{subsec:diff_3d_methods}. Finally, we compare the runtime of CoFF with baseline methods in Section \ref{subsec:runtime} \change{and discuss the limitations of the proposed 3D point cloud registration method in Section \ref{sub:limitations}.}

\subsection{Further analysis in geometrically ambiguous cases}
\label{subsec:factors}

\begin{figure}[tb]
  \centering
  \includegraphics[width=1.0\linewidth]{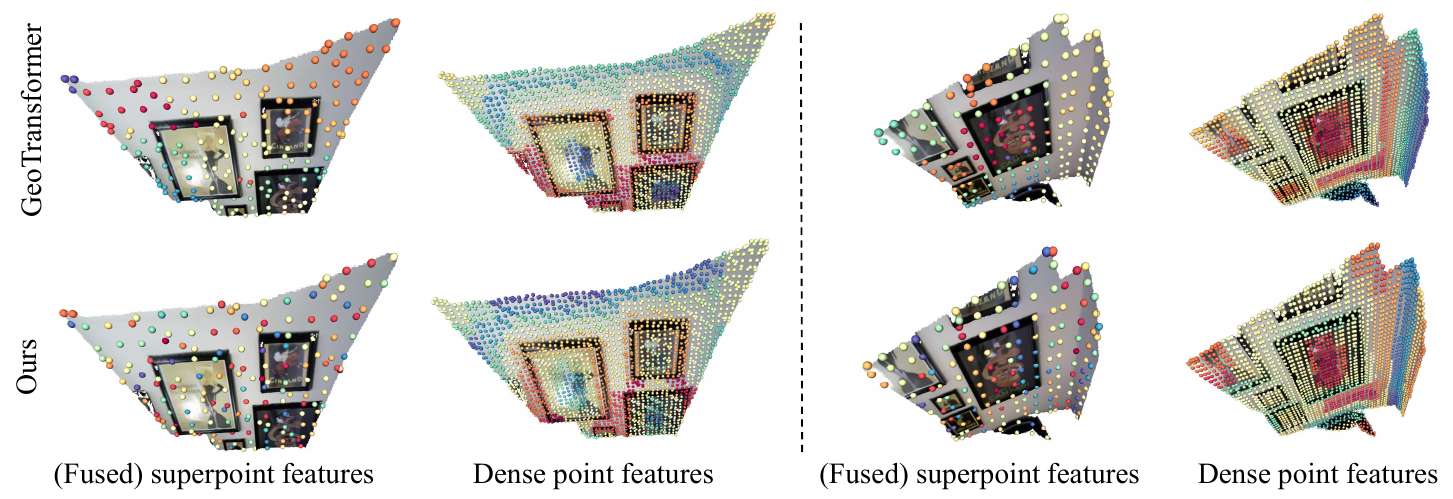}
  \caption{Point features overlaid on associated images. The features are mapped to a scalar space using t-SNE~\cite{van2008visualizing} and colorized with the spectral color map. Superpoint features learned by GeoTransformer~\cite{qin2022geometric} exhibit uniformly coded colors in planar overlap regions. In contrast, features generated by our method from the fusion of superpoint features with patch-wise image features display a richer color variety in the overlap region.
  }
  \label{fig:features}
\end{figure}

\begin{figure}[tb]
  \centering
  \includegraphics[width=1.0\linewidth]{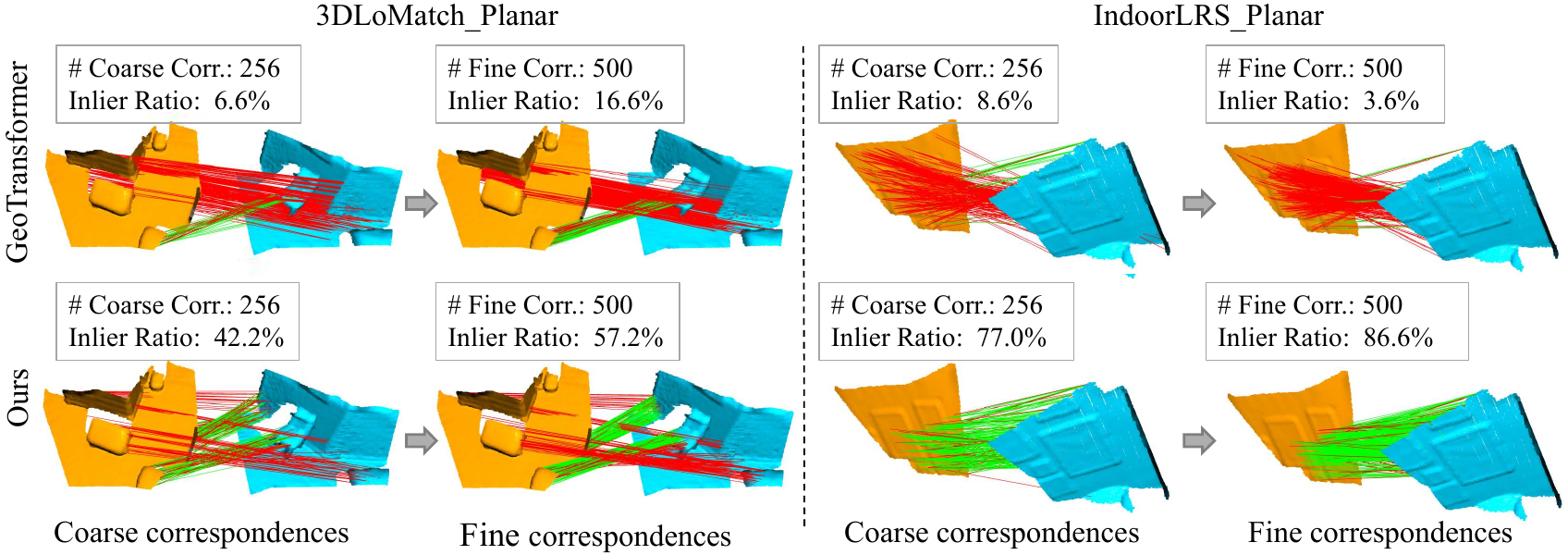}
  \caption{Coarse and fine correspondences on selected subset datasets. GeoTransformer~\cite{qin2022geometric} struggles with establishing reliable coarse correspondences in presence of ambiguous geometry, achieving low inlier ratios of only 6.6\% and 8.6\%. In contrast, our method, utilizing patch-wise image features, significantly improves the reliability of coarse correspondence, with inlier ratios of 42.2\% and 77.0\%.
  }
  \label{fig:inlier_ratio}
\end{figure}

In \cref{fig:features}, we demonstrate the distinctiveness of learned features. We observe that CoFF superpoint features, which are used for coarse matching in the coarse-to-fine matching process, exhibit notable distinctiveness for points at the borders of scenes. This distinctiveness arises from differences in their local image patches, where only superpoints with identical image patches (\eg, 64x64 in size) exhibit identical features. Furthermore, focusing on points within the overlap region between pairs (\cref{fig:features}, left \& right), we find that CoFF superpoint features align more closely with the color patterns of the background and exhibit local variability. This indicates a stronger capability to distinguish dense points in geometrically planar regions, enhancing the robustness and accuracy of the registration process.

In addition, we visualize the quality of established coarse and fine correspondences on the subset datasets in \cref{fig:inlier_ratio}. These subset datasets are particularly challenging because features learned purely from geometry can be ambiguous. Methods such as GeoTransformer~\cite{qin2022geometric}, which learn superpoint features based solely on geometry, often establish less reliable correspondences during coarse matching (6.6\% and 8.6\%). In contrast, our learnt superpoint features, which are fused with patch-wise image features, establish a significantly higher number of reliable coarse correspondences (42.2\% and 77.0\%), demonstrating its effectiveness in alleviating coarse matching ambiguity. Consequently, benefiting from both these reliable coarse correspondences and the features enhanced by pixel-wise image features, our method achieves a much higher inlier ratio (57.2\% and 86.6\%) for fine correspondences compared to the baseline GeoTransformer~\cite{qin2022geometric}, which achieves only 16.6\% and 3.6\% inlier ratios of fine correspondences. During coarse-to-fine matching, the high quality of our fine correspondences ensures a higher probability of accurate transformation estimation, thereby enhancing overall registration performance.

\subsection{Effectiveness of different image features}
\label{subsec:differ_img_feats}

\begin{table}[h]
    \caption{Results of using different features:
  3D point cloud features ($\mathbf{F}^\mathrm{3D}$ + $\mathbf{f}^\mathrm{3D}$), pixel-wise image features ($\mathbf{F}^\mathrm{2D}$), and patch-wise image features ($\mathbf{f}^\mathrm{2D}$).}
      \vspace{0.5em}
  \label{tab:abla_modules}
  \centering
  \resizebox{0.99\columnwidth}{!}{
  \begin{tabular}{ccc|ccc|ccc}
    \toprule
    \multirow{2}{*}{$\mathbf{F}^\mathrm{3D}$+$\mathbf{f}^\mathrm{3D}$}~~&~~\multirow{2}{*}{$\mathbf{F}^\mathrm{2D}$}~~&~~\multirow{2}{*}{$\mathbf{f}^\mathrm{2D}$}~& \multicolumn{3}{c|}{3DMatch} & \multicolumn{3}{c}{3DLoMatch} \\
    ~&~&~&~FMR(\%)$\uparrow$~&~IR(\%)$\uparrow$~&~RR(\%)$\uparrow$~&~FMR(\%)$\uparrow$~&~IR(\%)$\uparrow$~&~RR(\%)$\uparrow$~\\
    \midrule
    \checkmark~&&&~97.9~&~71.9~&~92.0~~~&~~88.3~&~43.5~&~74.0~\\
    \checkmark&\checkmark~&~&~\underline{98.8}~&~\textBF{76.2}~&~\underline{94.2}~~~&~~90.1~&~\textBF{48.2}~&~\underline{78.0}~\\
    \checkmark~&~&\checkmark~&~98.3~&~68.7~&~91.5~~~&~~\underline{92.7}~&~40.5~&~77.8~\\
    \checkmark&\checkmark&\checkmark&~\textBF{99.5}~&~\underline{74.3}~&~\textBF{95.9}~~~&~~\textBF{93.6}~&~\underline{47.4}~&~\textBF{81.6}~\\
  \bottomrule
  \end{tabular}
  }
\end{table}


We show the evaluation results of our feature fusion method in \cref{tab:abla_modules}. Compared to our baseline without image features, the inclusion of pixel-wise image features leads to significant improvements, with FMR by 0.9\%, 1.8\% and RR by 2.2\%, 4.0\% on 3DMatch~\cite{dai20183dmv} and 3DLoMatch~\cite{predator2021}, respectively.
When fusing only patch-wise image features with point cloud features, performance improves on 3DLoMatch \cite{predator2021} but slightly decreases on 3DMatch~\cite{zeng20163dmatch}. This could be attributed to the distraction caused by patch-wise image features in scenarios rich in geometric details. 
\change{Using the complete feature fusion module yields the best performance, and the results corroborate that pixel-wise and patch-wise image features are complementary, not redundant.}
This approach effectively balances their contributions, resulting in the highest FMR and RR across both datasets.

\subsection{\change{Impact of number of RGB images}}
\label{sec:sup_num_images}

\begin{table}[h]
  \caption{Quantitative results on numbers of images used.}
  \vspace{0.5em}
  \label{tab:num_imgs}
  \centering
    \resizebox{0.99\columnwidth}{!}{
  \begin{tabular}{c|ccc|ccc}
    \toprule
     & \multicolumn{3}{c|}{3DMatch}&  \multicolumn{3}{c}{3DLoMatch}\\
    \midrule
    Num. of Images & FMR (\%)$\uparrow$ & IR (\%)$\uparrow$ & RR (\%)$\uparrow$ & FMR (\%)$\uparrow$ & IR (\%)$\uparrow$ & RR (\%)$\uparrow$ \\
    \midrule
    \change{0}~&~\change{97.9}~&~\change{71.9}~&~\change{92.0}~~~&~~\change{88.3}~&~\change{43.5}~&~\change{74.0}~\\
    1~&~99.1~&~71.4~&~92.5~~~&~~92.9~&~43.5~&~79.7~\\
    2~&~\underline{99.4}~&~73.7~&~\underline{93.9}~~~&~~\textBF{94.0}~&~45.8~&~\underline{81.3}~\\   
    3~&~\textBF{99.5}~&~\textBF{74.3}~&~\textBF{95.9}~~~&~~\underline{93.6}~&~\textBF{47.4}~&~\textBF{81.6}~\\
    4~&~99.2~&~73.9~&~93.5~~~&~~92.9~&~46.6~&~79.7~\\
    5~&~99.3~&~\underline{74.0}~&~93.7~~~&~~93.0~&~\underline{47.3}~&~79.9~\\
  \bottomrule
  \end{tabular}
  }
\end{table}

\begin{figure}[tb]
  \centering
  \includegraphics[width=0.7\linewidth]{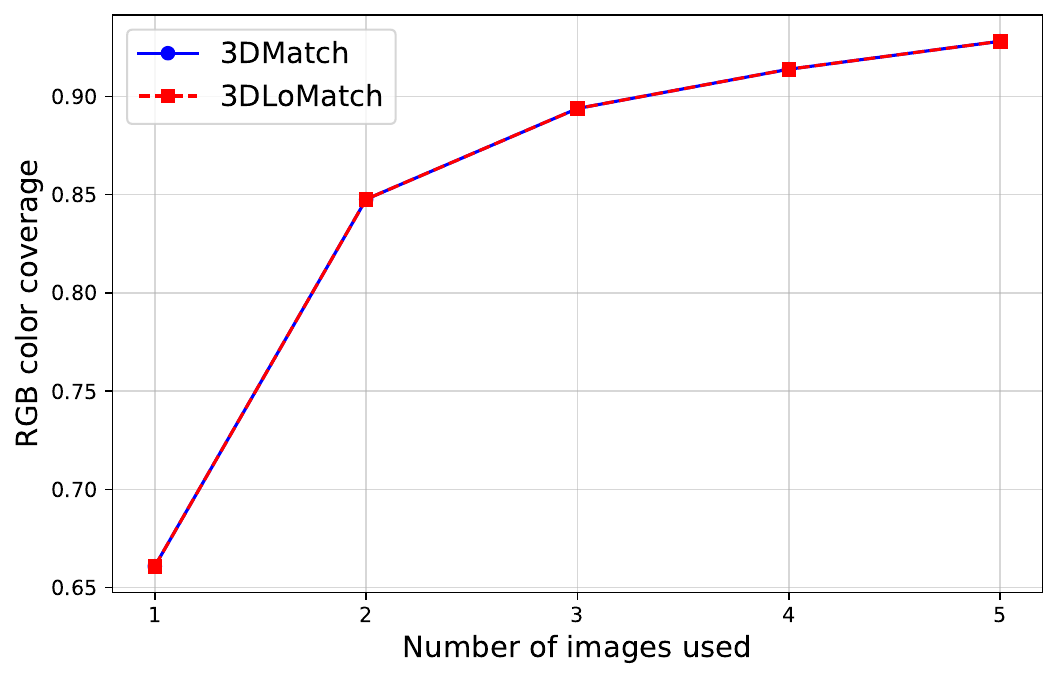}
  \caption{\change{RGB color coverage for point clouds in relation to number of images.}}
  \label{fig:missing_ratio_3dmatch}
\end{figure}

\change{The resolution of the data can impact the extraction of image features, ultimately affecting the registration performance. In our case, when the image resolution exceeds the point cloud resolution, 
the pixel closest
to a given point is selected, and its associated 
image features are assigned to the point. Conversely, when the image resolution is lower than the point cloud resolution or when occlusions cause the image coverage to be lower than the point cloud coverage, some points may lack valid corresponding pixels. 
In such cases, these points are still processed by the approach, but without any associated image features.
To analyze this effect, we ablate 
by varying the number of input images.}

\cref{tab:num_imgs} shows that using one to three RGB images enhances the RR for both 3DMatch~\cite{zeng20163dmatch} and 3DLoMatch~\cite{predator2021} datasets, with RR increasing from 92.5\% to 95.9\% for 3DMatch~\cite{zeng20163dmatch} and 79.9\% to 81.6\% for 3DLoMatch~\cite{predator2021}. This improvement highlights the benefits of using extra images in improving registration quality, \change{as more points are covered by RGB colors (\cf \cref{fig:missing_ratio_3dmatch})}. However, the optimal performance is achieved with three images; beyond this, the performance drops, likely due to noise from additional image views. Therefore, we select three images for image feature extraction on all four datasets.

\subsection{Impact of multi-view feature selection} 
\label{subsec:multiview_feature}

In \cref{tab:abla_multiview}, we compare different multi-view image feature selection strategies for selecting the pixel-wise image feature from multiple views. \emph{Random} denotes randomly assigning one of the pixel-wise image features to the 3D point. \emph{Complement} denotes assigning the pixel-wise image feature complementarily: starting with the feature vector extracted from the closest image, when the pixel corresponding to the 3D point is missing in the current image, we proceed to the next closest image. This allows us to use image features from the same image as much as possible. \emph{Mean} indicates assigning the average of multiples image features to the 3D point. As shown in \cref{tab:abla_multiview}, the \emph{Complement} strategy yields the best results in terms of FMR, IR, and RR metrics on both the 3DMatch~\cite{zeng20163dmatch} and 3DLoMatch~\cite{predator2021} datasets. We attribute this improvement to the fact that the \emph{Complement} strategy can ensure the overall consistency of the input features. 

\begin{table}[h]
    \caption{Results of different multi-view image feature selection strategies.}
    \vspace{0.5em}
  \label{tab:abla_multiview}
  \centering
  \resizebox{0.99\columnwidth}{!}{
  \begin{tabular}{c|ccc|ccc}
    \toprule
    & \multicolumn{3}{c|}{3DMatch}&  \multicolumn{3}{c}{3DLoMatch} \\
    ~&~FMR(\%)$\uparrow$~&~IR(\%)$\uparrow$~&~RR(\%)$\uparrow$~&~FMR(\%)$\uparrow$~&~IR(\%)$\uparrow$~&~RR(\%)$\uparrow$~\\
    \midrule
    Random~&~\underline{99.4}~&~71.2~&~93.2~~&~~92.2~&~43.1~&~78.0~\\
    Mean~&~99.0~&~\underline{73.8}~&~\underline{94.7}~~&~~\underline{93.2}~&~\underline{46.2}~&~\underline{80.6}~\\
    Complement~&~\textBF{99.5}~&~\textBF{74.3}~&~\textBF{95.9}~~&~~\textBF{93.6}~&~\textBF{47.4}~&~\textBF{81.6}~\\
  \bottomrule
  \end{tabular}
  }
\end{table}

\subsection{Incorporation with different 3D registration methods}
\label{subsec:diff_3d_methods}

\begin{table}[h]
    \caption{Registration results of RR with different 3D methods. The registration results after incorporating CoFF are reported, with improvements compared to the original 3D methods reported in parentheses on specific datasets.}
    \vspace{0.5em}
  \label{tab:sub_regi_methods}
  \centering
  \resizebox{0.99\columnwidth}{!}{
  \begin{tabular}{lc|cccc}
   \toprule
    Fusion & 3D method & 3DMatch & 3DLoMatch & 3DMatch\_Planar & 3DLoMatch\_Planar \\
    \midrule
    CoFF~&~Predator~&~89.3 (+0.3)~&~61.2 (+3.1)~&~61.5 (+ 6.5)~&~49.3 (+ 1.4)~\\
    CoFF~&~GeoTransformer~&~95.9 (+4.4)~&~81.6 (+7.6)~&~90.5 (+14.1)~&~70.4 (+10.9)~\\
  \bottomrule
  \end{tabular}
  }
\end{table}

We incorporate CoFF into 3D registration methods with or without coarse-to-fine matching, presenting results on 3DMatch / 3DLoMatch and their subsets in~\cref{tab:sub_regi_methods}. CoFF is compatible with both settings. When combining CoFF with Predator~\cite{predator2021}, we assign pixel-wise image features to input point clouds and incorporate features of coarsest points with patch-wise image features. CoFF shows larger improvement with GeoTransformer~\cite{qin2022geometric} compared to Predator~\cite{predator2021}, indicating that our two-stage feature fusion approach is better suited for coarse-to-fine matching, specifically in mitigating ambiguity during coarse matching.

\subsection{Runtime analysis}
\label{subsec:runtime}

\begin{table}[h]
  \caption{Runtime comparison with different point cloud registration methods. \change{The comparison is done on 3DMatch test set, and the average time of 1,623 pairs are reported.}}
  \vspace{0.5em}
  \label{tab:runtime}
  \centering
    \resizebox{0.99\columnwidth}{!}{
  \begin{tabular}{cc|ccc}
    \toprule
    Method & \change{3D backbone type} & Data (s)$\downarrow$ & Model (s)$\downarrow$ & Total (s)$\downarrow$ \\
    \midrule
    
    FCGF~\cite{FCGF2019}~&~\change{Minkowski}~&~0.368~&~0.085~&~0.453~\\
    Predator~\cite{predator2021}~&~\change{KPConv}~&~0.431~&~\textBF{0.026}~&~0.457~\\
    CoFiNet~\cite{yu2021cofinet}~&~\change{KPConv}~&~\textBF{0.013}&~0.125~&~\textBF{0.138}~\\    
    PCR-CG~\cite{zhang2022pcr}~&~\change{KPConv}~&~0.513~&~0.048~&~0.561~\\
    GeoTransformer~\cite{qin2022geometric}~&~\change{KPConv}~&~0.071~&~0.076~&~0.147~\\    
    CoFF(ours)~&~\change{KPConv}~&~0.098~&~0.244~&~0.342~\\
  \bottomrule
  \end{tabular}
  }
\end{table}

In addition to evaluating effectiveness, we also assess the efficiency of our method and compare it to other point cloud registration approaches. \change{Similar to other existing methods~\cite{predator2021,yu2021cofinet,zhang2022pcr,qin2022geometric}, our method has an overall time complexity of $O(n^2)$ due to the cross-attention module,
where $n$ represents the number of superpoints (\ie, points at the sparsest level).}
Given that CoFF integrates extra image features, it results in a larger model size compared to purely geometry-based methods for point cloud registration.
However, we mitigate training burden by offline extracting pixel-wise image features, and leveraging pre-trained image models that enable CoFF to converge in fewer epochs. Furthermore, we provide a runtime comparison during inference in \cref{tab:runtime}. To make a fair comparison, we evaluate all methods on the 3DMatch benchmark dataset and execute them on a single GeForce RTX 3090 Ti with an AMD Ryzen 7 5800X 8-Core Processor. 
We notice that both CoFiNet~\cite{yu2021cofinet} and GeoTransformer~\cite{qin2022geometric}, along with our method, demonstrate superior data loading efficiency compared to other methods. Although our approach entails longer total runtime than CoFiNet~\cite{yu2021cofinet} and GeoTransformer~\cite{qin2022geometric} due to additional image feature extraction, it surpasses PCR-CG~\cite{zhang2022pcr} in efficiency, which also incorporates image feature extraction. 

\subsection{\change{Limitations}}
\label{sub:limitations}

Our work has several limitations. 
First, our approach relies on the availability of RGB images. While this enables us to leverage numerous pre-trained models on 2D image datasets, it also limits the applicability when only colorized point clouds are available. 
Second, we assume that accurate co-registration between point clouds and associated RGB images is available. This might not hold in all scenarios and could lead to degraded performance in case of inaccurate co-registration. However, 
\change{the co-registration can be handled as part of the data preprocessing step. We implemented a validation step during preprocessing, where we compare 3D points with their projections onto the images. If significant misalignment is detected, \change{\eg, via a reprojection error threshold or a dedicated 2D-3D alignment check~\cite{assess_align}}, we recommend correcting it using either camera-scanner calibration methods~\cite{calibrate18,calibrate23} with a checkerboard or} recent deep learning-based methods, \eg, 2D3D-MATR~\cite{li20232d3d} and FreeReg~\cite{wang2024freereg}. 
Lastly, in scenarios where both geometry and color information are ambiguous, our method may struggle to identify distinctive features, potentially leading to registration failure (\cf failure cases in \ref{sec:sup_more_qualitative}). This limitation is not unique to our method but is inherent to methods relying solely on geometry and color data. Effective solutions in such scenarios typically require additional information, \eg, sensor data encoding positional and orientational  changes between data acquisition locations. Alternatively, integrating 
semantic prior knowledge (\eg, from pre-trained language-image CLIP~\cite{radford2021learning} models) may enhance the robustness of our method. 
\section{Conclusion}
\label{sec:conclusion}

In this paper, we present CoFF, a cross-modal feature fusion method for pairwise point cloud registration. To address the challenges posed by ambiguous geometry, CoFF incorporates 3D point cloud features with both pixel-wise and patch-wise image features through a two-stage fusion process. This involves assigning pixel-wise image features to 3D input point clouds to enhance learned 3D features, and integrating patch-wise image features with superpoint features to alleviate coarse matching ambiguity. Benefiting from using both types of image features, our method shows superior performance for the point cloud registration task on our subset datasets that mainly feature the scenarios with geometrically ambiguous in the overlap region. In addition, our method can effectively improve the overall registration performance on different scenarios, not limited to geometrically ambiguous cases. It sets new state-of-the-art registration recall on all common datasets, including 3DMatch, 3DLoMatch, IndoorLRS and ScanNet++.
\change{In future work, it would be interesting to explore the joint estimation of image-point cloud and point cloud-point cloud transformations.}

\section*{CRediT authorship contribution statement}

\textbf{Zhaoyi Wang}: Methodology, Data curation, Investigation, Formal analysis, Validation, Visualization, Writing – original draft, Writing – review \& editing, Software.
\textbf{Shengyu Huang}: Conceptualization, Methodology, Formal Analysis, Validation, Visualization, Writing – review \& editing.
\textbf{Jemil Avers Butt}: Conceptualization, Methodology, Formal analysis, Writing – review \& editing.
\textbf{Yuanzhou Cai}: Data curation, Writing – review \& editing, Software.
\textbf{Matej Varga}: Writing – review \& editing, Supervision.
\textbf{Andreas Wieser}: Conceptualization, Writing – review \& editing, Resources, Supervision, Project administration, Funding acquisition.



\section*{Declaration of competing interest}

The authors declare that they have no known competing financial interests or personal relationships that could have appeared to influence the work reported in this paper.

\section*{Acknowledgments}

The first author was supported by the China Scholarship Council (CSC).

\section*{\change{Use of Generative AI and AI-assisted technologies in the writing process}}

\change{
During the preparation of this work, the authors used ChatGPT to assist in refining the readability and language of the manuscript. The authors then reviewed and edited the resulting text and take full responsibility for the content of the entire publication.}

\appendix

\section{Loss functions}
\label{sec:sup_loss}

\paragraph{\textnormal{\textbf{Superpoint feature matching loss}}}

We use the overlap-aware circle loss defined in~\cite{qin2022geometric,yu2023rotation}, where the overlap ratio between superpoint patches is used to weight different ground-truth superpoint correspondences. Specifically, for each superpoint $\mathbf{p}^\mathrm{3D}_i \in \mathbf{P}^\mathrm{3D}$ with associated feature $\mathbf{f}^\mathrm{3D}_{\mathbf{P},i} \in \mathbf{f}^\mathrm{3D}_\mathbf{P}$ (after normalization), we sample a positive set of superpoints from $\mathbf{Q}$, denoted as $\zeta^i_p$, when the overlap ratio is above $10\%$. We also sample a negative set of superpoints from $\mathbf{Q}$, denoted as $\zeta^i_n$, when the overlap ratio is $0$.

The positive weight $\beta^{i,j}_p = \gamma(d^j_i - \mathrm{\Delta}_p)$ and negative weight $\beta^{i,k}_n = \gamma(\mathrm{\Delta}_n - d^j_i)$ are computed for each sample individually, with the positive margin $\mathrm{\Delta}_p=0.1$, negative margin $\mathrm{\Delta}_n=1.4$, and a hyper-parameter $\gamma=24$ by default. Then, the superpoint feature matching loss for $\mathbf{P}$ is computed as:


\begin{multline}
\label{eq:superpoint_loss}
\mathcal{L}^{\mathcal{P}}_\mathrm{point} = \frac{1}{n_c} \sum_{i=1}^{n_c} \log(1 + \\
\sum_{\mathbf{G}^{\mathcal{Q}}_j \in \zeta^i_p} e^{\lambda^j_i\beta^{i,j}_p (d^j_i - \mathrm{\Delta}_p)} \sum_{\mathbf{G}^{\mathcal{Q}}_k \in \zeta^i_n} e^{\beta^{i,k}_n (\mathrm{\Delta}_n - d^k_i)}),
\end{multline}
where $n_c$ denotes the number of superpoint patches in $\mathbf{P}$ which have at least one positive patch in $\mathbf{Q}$, $\lambda^j_i$ denotes the overlap ratio between $\mathbf{G}^{\mathcal{P}}_i$ and $\mathbf{G}^{\mathcal{Q}}_j$, and $d^j_i$ denotes the $l^2$-norm distance between superpoint features.

The superpoint feature matching loss for $\mathbf{Q}$ is computed in the same way, and the final superpoint feature matching loss is $\mathcal{L}_\mathrm{point} = (\mathcal{L}^{\mathbf{P}}_\mathrm{point} + \mathcal{L}^{\mathbf{Q}}_\mathrm{point})/2$.

\paragraph{\textnormal{\textbf{Patch-wise feature matching loss}}}

The patch-wise feature matching loss, denoted as $\mathcal{L}^{\mathrm{patch}}_\mathrm{img}$, is calculated in a manner similar to superpoint feature matching loss. We simply replace the superpoint features $\mathbf{f}^\mathrm{3D}_\mathbf{P}$ and $\mathbf{f}^\mathrm{3D}_\mathbf{Q}$ (after normalization) with the corresponding patch-wise image features $\mathbf{f}^\mathrm{2D}_\mathbf{P}$ and $\mathbf{f}^\mathrm{2D}_\mathbf{Q}$ for computing the $l^2$-norm distance.

\paragraph{\textnormal{\textbf{Dense point matching loss}}} 

We adopt a negative log-likelihood loss on the assignment matrix $\mathbf{S}_l$ to supervise dense point correspondences. Specifically, for each superpoint correspondence, a set of indices $\mathbf{\Omega}^{\mathrm{f}}_i$ related to matched dense points within two matched patches is extracted with a matching radius $\mathrm{\Delta}_r$ ($\mathrm{\Delta}_r$ = 0.05 by default). The sets of unmatched points in the two superpoint patches are denotes as $\mathbf{\Phi}_i$ and $\mathbf{\Psi}_i$. The individual dense point matching loss is computed as:

\begin{multline}
\label{eq:dense_point_loss}
\mathcal{L}_{\mathbf{\Omega}^\mathrm{f}, i} = -\sum_{{(x, y) \in \mathbf{\Omega}^{\mathrm{f}}_i}} \log {s}^l_{x, y} - \sum_{x \in \mathbf{\Phi}_i} \log {s}^l_{x, |\mathbf{G}^{\mathbf{Q}}_j|+1} \\
- \sum_{y \in \mathbf{\Psi}_i} \log {s}^l_{|\mathbf{G}^{\mathbf{P}}_i|+1, y},
\end{multline}
where ${s}^l_{x, y}$ denotes the entry on the row $x$ and column $y$ of $\mathbf{S}_l$.

The final dense point matching loss $\mathcal{L}_{\mathbf{\Omega}, i}$ is computed by averaging the individual loss over all sampled superpoint matches: $\mathcal{L}_{\mathbf{\Omega}^\mathrm{f}} = \frac{1}{|\mathbf{\Omega}^\mathrm{f}|} \sum^{|\mathbf{\Omega}^\mathrm{f}|}_{i=1} \mathcal{L}_{\mathbf{\Omega}^\mathrm{f}, i}$.

\section{\change{Results with different thresholds}}
\label{sec:sup_differ_thresholds}

\begin{figure*}[tb]
  \centering
  \includegraphics[width=0.7\linewidth]{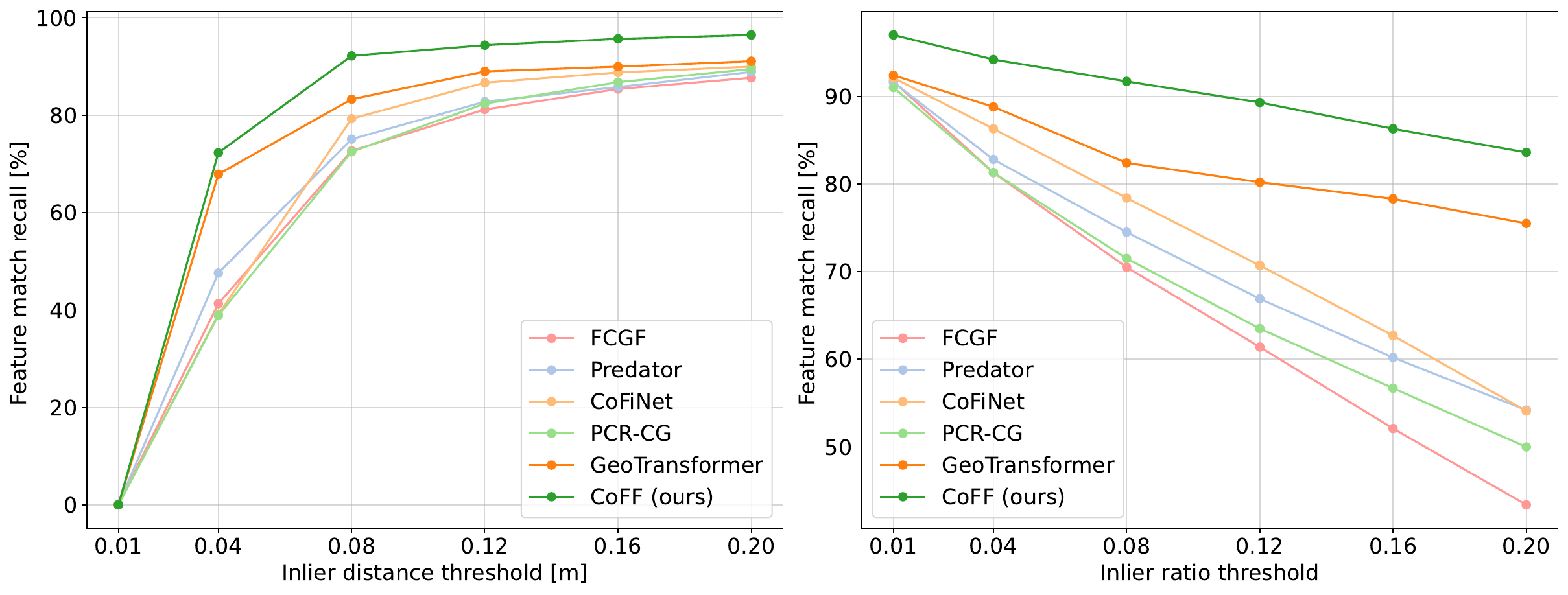}
  \caption{\change{Feature matching recall in relation to inlier distance threshold $\mathrm{\Delta}_r$ (left) and inlier ratio threshold $\tau_2\%$ (right).}}
  \label{fig:fmr_vs_threshold}
\end{figure*}

\begin{figure}[tb]
  \centering
  \includegraphics[width=0.7\linewidth]{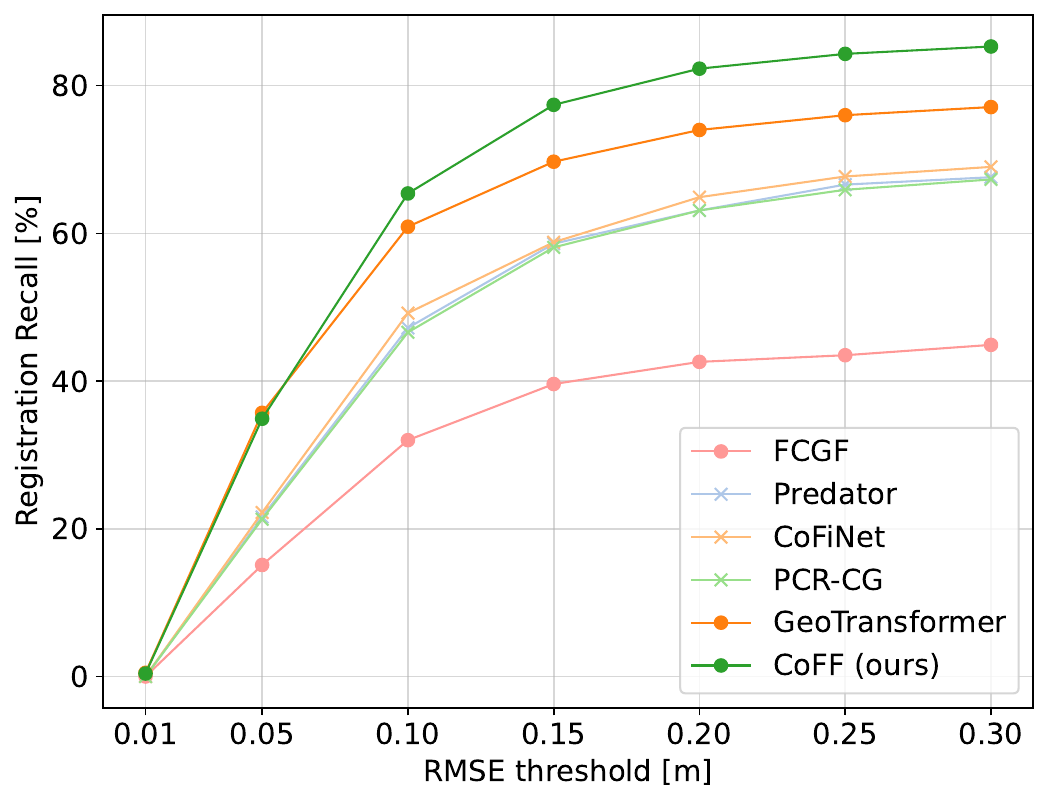}
  \caption{\change{Registration recall in relation to RMSE threshold.}}
  \label{fig:rr_vs_rmse}
\end{figure}

\change{We further evaluate the performance of different methods over a wide range for the allowable inlier distances, minimum inlier ratios, and minimum RMSEs. As shown in~\cref{fig:fmr_vs_threshold} and \cref{fig:rr_vs_rmse}, our method achieves the best FMR and RR under a wide range of different thresholds.}

\section{\change{Results on outdoor datasets}}
\label{sec:outdoor}

\change{We further evaluate our method on the ETH3D~\cite{eth3d} dataset, where we select four outdoor scenes---courtyard, facade, playground, and terrace---for training and testing. These outdoor scenes pose greater challenges than indoor scenes due to their large scale and the influence of environmental factors (\eg, natural illumination conditions). The RGB images are captured using a Nikon D3X camera, while the point clouds are collected with a Faro Focus X 330 scanner. To generate sufficient training samples, depth images are rendered from the point clouds, each corresponding to an RGB image~\cite{eth3d}. We preprocess the data to adapt it for the point cloud registration task. Specifically, we undistort the RGB and depth images using COLMAP~\cite{schoenberger2016sfm,schoenberger2016mvs}. Each point cloud is then reconstructed from an RGB image and its corresponding depth image using TSDF volumetric fusion. We select point cloud pairs with at least 30\% overlap, resulting in 217 training pairs and 32 test pairs. For comparison, we select GeoTransformer~\cite{qin2022geometric} as the strongest baseline. During training, both our method and GeoTransformer use a voxel size of 0.1 m.
The results demonstrate that our method outperforms GeoTransformer (71.9\% vs. 68.8\% of RR), 
although we believe further investigation needs to be performed.
For even larger-scale datasets, we recommend increasing the voxel size to reduce memory consumption and employing a simple engineering strategy—splitting the data into tiles—as commonly used in landslide monitoring~\cite{f2s3_v2} and semantic segmentation~\cite{robert2023spt}.
}

\section{More qualitative results}
\label{sec:sup_more_qualitative}

\begin{figure*}[t]
  \centering
  \includegraphics[width=0.95\linewidth]{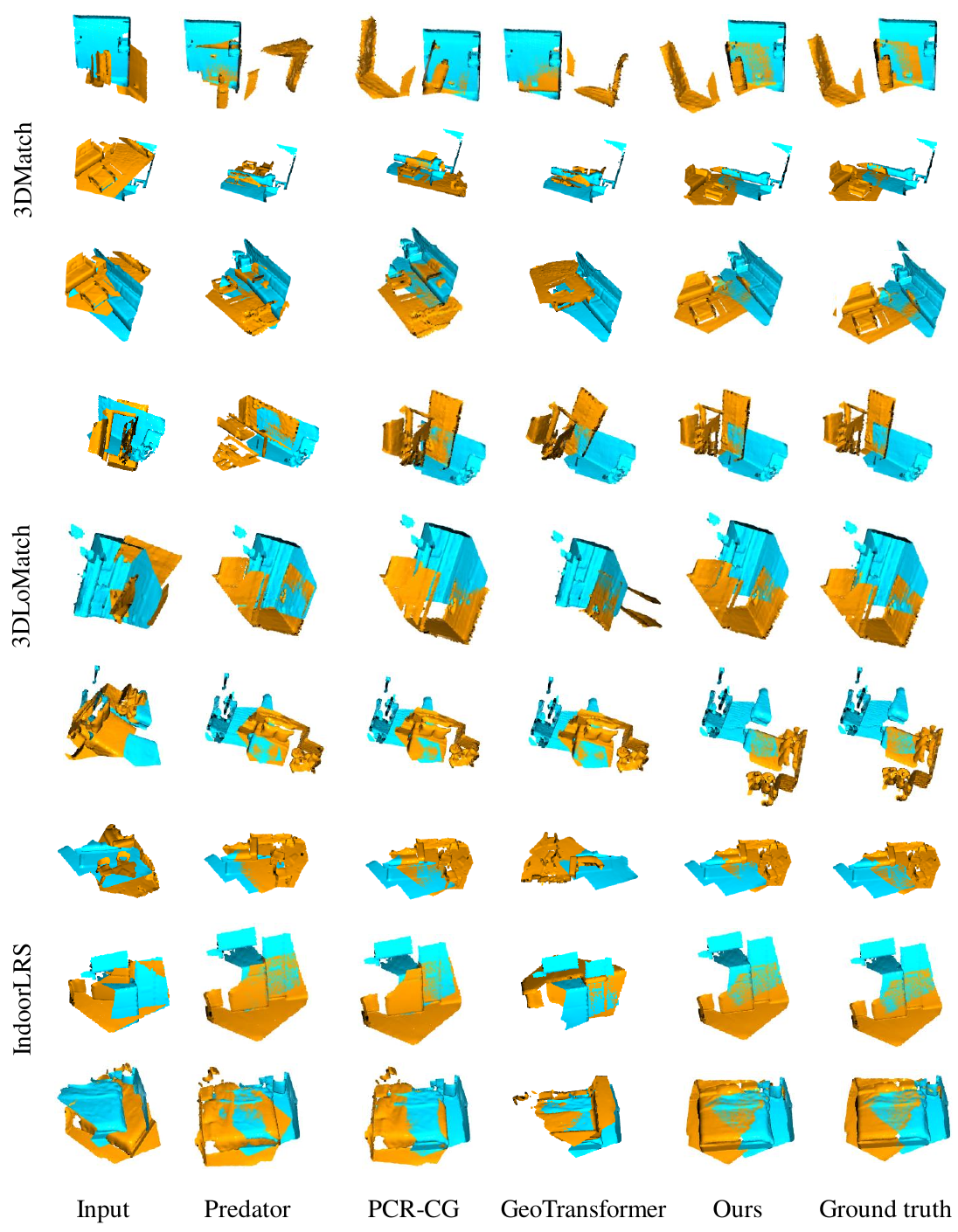}
  \vspace{-1.5em}
  \caption{More qualitative results on different datasets.
  }
  \label{fig:more_qualitative_results}
  \vspace{-2em}
\end{figure*}

\begin{figure*}[tb]
  \centering
  \includegraphics[width=1.0\linewidth]{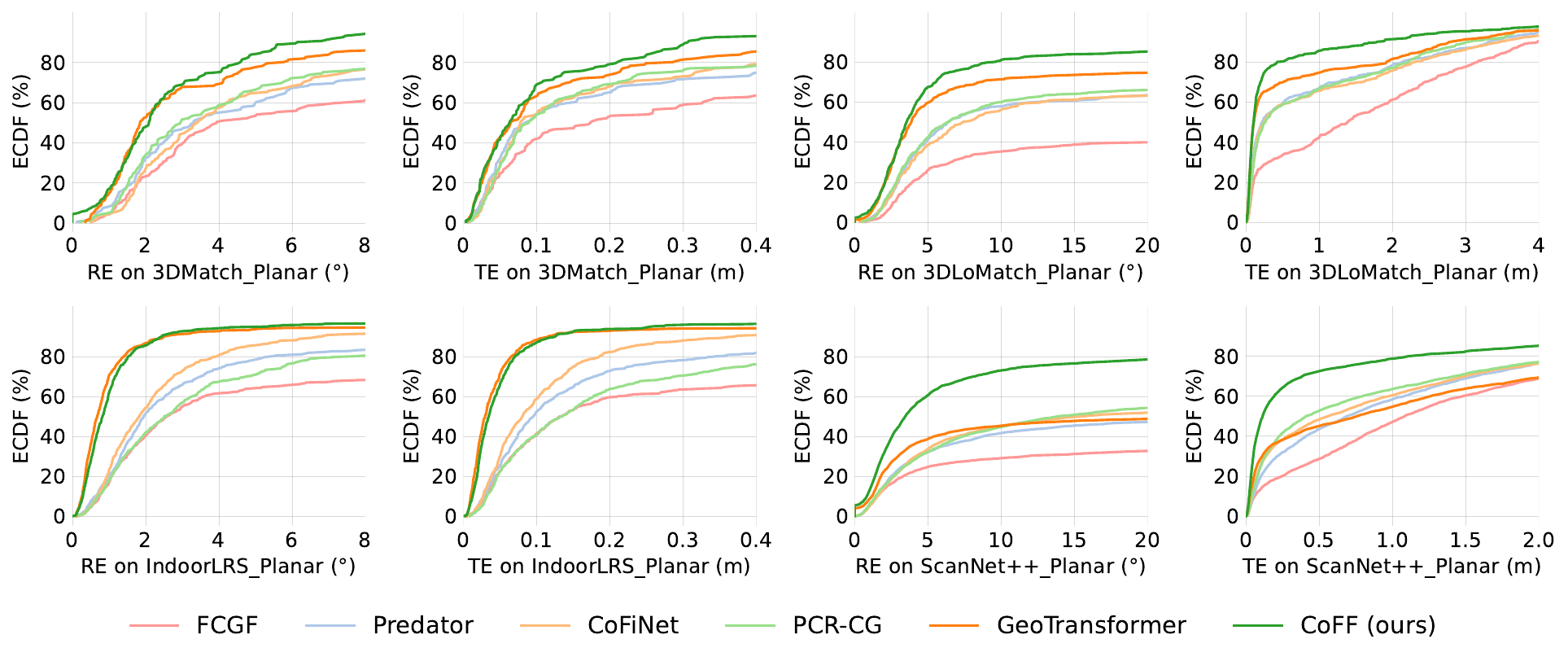}
  \caption{Rotation Error (RE) and Translation Error (TE) for all point cloud pairs on four subset datasets. To enhance visualization, we apply different x-axis value ranges for each of the four datasets based on their specific characteristics.}
  \label{fig:re_te_subset}
\end{figure*}

In \cref{fig:more_qualitative_results}, we \change{present qualitative results for different methods, where the examples are specifically selected to highlight most geometric ambiguity cases where our method succeeds while others fail.} We also present the RE and TE results on four subset datasets in \cref{fig:re_te_subset}.

\begin{figure*}[htb!]
  \centering
  \includegraphics[width=1.0\linewidth]{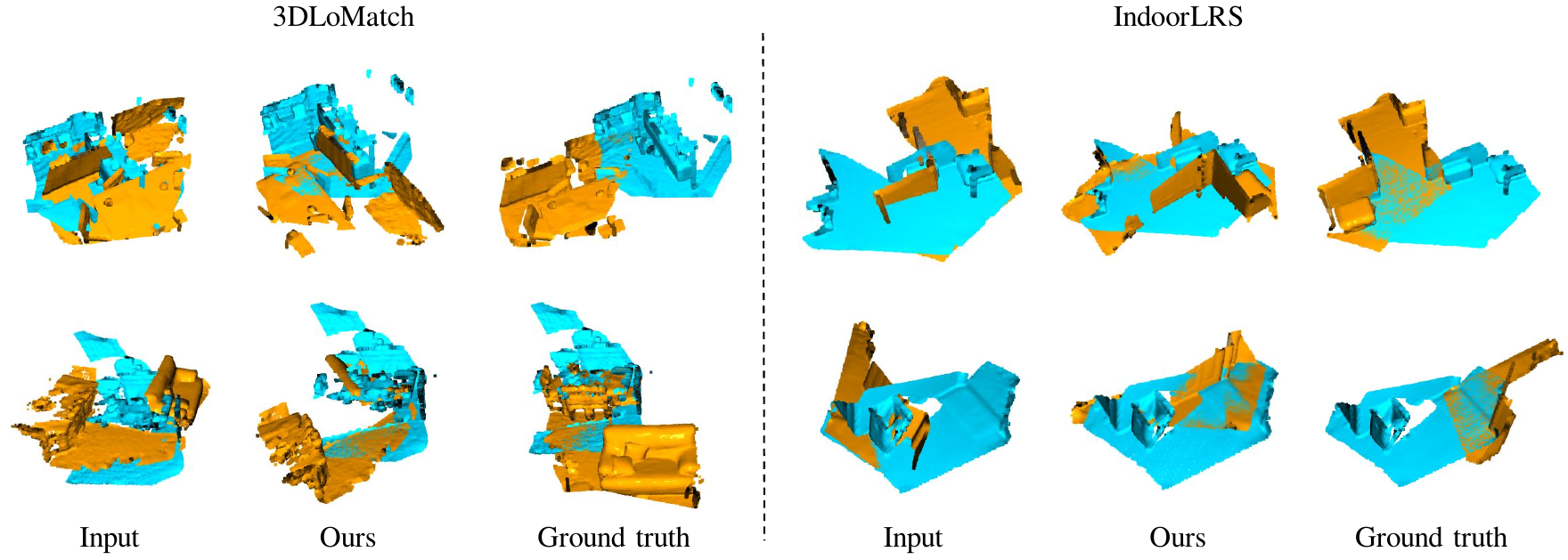}
  \caption{Failure cases on 3DLoMatch and IndoorLRS datasets, where both geometry and color information are ambiguous.}
  \label{fig:fail_cases}
\end{figure*}

Furthermore, \cref{fig:fail_cases} illustrates scenarios where our method does not perform well on the 3DLoMatch~\cite{predator2021} and IndoorLRS~\cite{Park2017coloricp} datasets. Typically, failures occur when the overlap in a point cloud pair is ambiguous both geometrically and radiometrically, making it challenging for our method to identify distinct features in the overlapping region and establish reliable correspondences. Incorporating semantic information could be a potential solution for addressing this ambiguity challenge, which we leave for future work.

\bibliography{biblio}

\end{document}